%% file: neurips_2026.tex
\documentclass{article}

\PassOptionsToPackage{numbers, compress}{natbib}
\usepackage[main, final]{neurips_2026}

\input{command/math_commands}

\input{command/mycommand}

\makeatletter
\newcommand\blfootnote[1]{%
  \begingroup
    \renewcommand\thefootnote{}%
    \renewcommand\@makefntext[1]{\noindent##1}
    \footnote{#1}%
    \addtocounter{footnote}{-1}%
  \endgroup
}
\makeatother

\usepackage{hyperref}
\usepackage[utf8]{inputenc} 
\usepackage[T1]{fontenc}    
\usepackage{hyperref}       
\usepackage{url}            
\usepackage{booktabs}       
\usepackage{amsfonts}       
\usepackage{nicefrac}       
\usepackage[dvipsnames]{xcolor}         

\usepackage{anyfontsize}
\usepackage{graphicx}
\usepackage{amsmath}
\usepackage{amssymb}
\usepackage{booktabs}
\usepackage{amsthm}
\usepackage{algpseudocode}

\usepackage{color}
\usepackage{xcolor}
\usepackage{colortbl}
\usepackage[export]{adjustbox}
\usepackage{arydshln}
\usepackage{wrapfig}
\usepackage{enumitem}

\usepackage[capitalize]{cleveref}

\definecolor{LightCyan}{rgb}{0.88,1,1}
\definecolor{Blue}{rgb}{0, 0.5, 1}
\definecolor{Green}{rgb}{0.0, 0.8, 0.0 }
\definecolor{Red}{rgb}{0.95, 0.55, 0.6}
\definecolor{Skyblue}{rgb}{0.6, 0.6, 0.95 }

\NewDocumentEnvironment{suptitle}{ +b }{
    \onecolumn{#1}%
}{}

\NewDocumentCommand{\supptitle}{s}{
\begin{suptitle}
        \centering
        \rule{\textwidth}{0.03cm}\\[0.1cm]
        -Supplementary Material-\\[0.2cm]
        {\Large 
            \textbf{\mytitle }
        }\\
        \rule{\textwidth}{0.03cm}\\[0.2cm]
\end{suptitle}}

\theoremstyle{plain}
\newtheorem{lemma}{Lemma}[section]

\theoremstyle{definition}
\newtheorem{definition}[lemma]{Definition}

\theoremstyle{remark}

\title{\mytitle}

\author{%
{Jihwan Bang$^1$ \quad Sumyeong Ahn$^2$ \quad Hwanjun Song$^1$ \quad Jae-Gil Lee$^{1*}$}\\
{KAIST$^1$\hspace{4em}KENTECH$^2$}\\
{\tt\small {\{jihwan.bang, songhwanjun, jaegil\}@kaist.ac.kr, sumyeongahn@kentech.ac.kr}}
}

\begin{document}

\maketitle

\renewcommand{\thefootnote}{\fnsymbol{footnote}} 
\footnotetext[1]{\hspace{0.3em} indicates the corresponding author.}

\input{revised_sections/00_abstract}
\input{revised_sections/01_Intro}
\input{revised_sections/02_Pre}

\input{revised_sections/03_methodology}

\input{revised_sections/04_exp}

\input{revised_sections/05_related}
\input{revised_sections/06_Conclusion}

\input{revised_sections/07_ACK}

\bibliographystyle{unsrt}
\bibliography{custom.bib}

\newpage
\appendix
\input{revised_sections/10_Appendix}


\newpage
\input{checklist.tex}

\end{document}

%% file: command/math_commands.tex
\usepackage{amsmath,amsfonts,bm}

\def\eqref#1{equation~\ref{#1}}

\def\1{\bm{1}}

\DeclareMathAlphabet{\mathsfit}{\encodingdefault}{\sfdefault}{m}{sl}
\SetMathAlphabet{\mathsfit}{bold}{\encodingdefault}{\sfdefault}{bx}{n}



%% file: command/mycommand.tex
\newcommand{\mytitle}{Training-Free Active Test-Time Adaptation for Vision-Language Models}
\newcommand{\alg}{\textbf{\texttt{ACTOR}}\xspace}

\usepackage{booktabs} 
\usepackage{bbm}
\usepackage{mathtools}

\usepackage{nccmath}
\usepackage{setspace}

\usepackage{caption}
\usepackage{subcaption}

\usepackage[linesnumbered,ruled,vlined]{algorithm2e}

\SetKwInput{KwInput}{Input}                
\SetKwInput{KwOutput}{Output}              

\SetCommentSty{mycommfont}

\SetAlCapSty{algcapsty}

\usepackage[T1]{fontenc}
\usepackage{wrapfig,lipsum,booktabs}

\usepackage{soul}
\usepackage{dsfont}
\usepackage{enumerate}
\usepackage{amsmath}
\usepackage{amsfonts}
\usepackage{bbm}
\usepackage{dsfont}
\usepackage[Symbol]{upgreek}
\usepackage{lscape}
\usepackage{caption}
\usepackage{balance}
\usepackage{xspace}
\usepackage{float}
\usepackage{kotex}

\usepackage{wasysym}

\usepackage{multirow}
\usepackage{array, boldline, rotating}

\usepackage{amssymb}
\usepackage{pifont}
\newcommand{\omark}{\textbf{$\mathcal{O}$}\xspace}%
\newcommand{\xmark}{\ding{55}\xspace}%

\renewcommand*\eqref[1]{(\ref{#1})}

\newcommand{\eg}{\emph{e.g.,~}}
\newcommand{\ie}{\emph{i.e.,~}}

\newcommand{\myparagraph}[1]{\vspace{0.07cm}\noindent\textbf{#1}~}



%% file: revised_sections/00_abstract.tex
\begin{abstract}
    Test-time adaptation (TTA) for vision-language models (VLMs) is usually studied in a fully unsupervised regime, where adaptation relies on pseudo-labels, confidence scores, or entropy computed by the model itself. This is a severe restriction on hard shifted samples: the model is asked to repair its own uncertain predictions without any external evidence. At the other extreme, supervised online adaptation assumes labels can directly train or correct the model. We study the intermediate deployment regime between these two extremes. In \emph{Active Test-time Adaptation for VLMs} (\textbf{Active VLM-TTA}), a pretrained VLM may spend an explicit labeling budget on selected test samples, but the prediction for the current sample must be emitted before its label can be used; queried labels are therefore costed, delayed evidence for future samples rather than free correction for the present one. Under this future-only constraint, the central question is how to reuse sparse labels conservatively. We instantiate this setting with \textbf{ACT}ive VLM-TTA for \textbf{O}nline \textbf{R}etrieval (\alg), a training-free plug-in module that routes uncertain samples among base prediction, label query, and memory-based correction using a consistency-and-reliability check over previously queried labels. Across ten cross-domain benchmarks and a domain-transfer benchmark based on ImageNet, \alg{} consistently improves strong VLM-TTA baselines on ResNet and ViT backbones with negligible inference overhead. Our implementation is publicly available at \url{https://github.com/kaist-dmlab/ACTOR}. 
\end{abstract}

%% file: revised_sections/01_Intro.tex
\section{Introduction}

\begin{wrapfigure}{r}{6.8cm}
    \vspace{-1.2cm}
    \centering
    \includegraphics[width=1.0\linewidth]{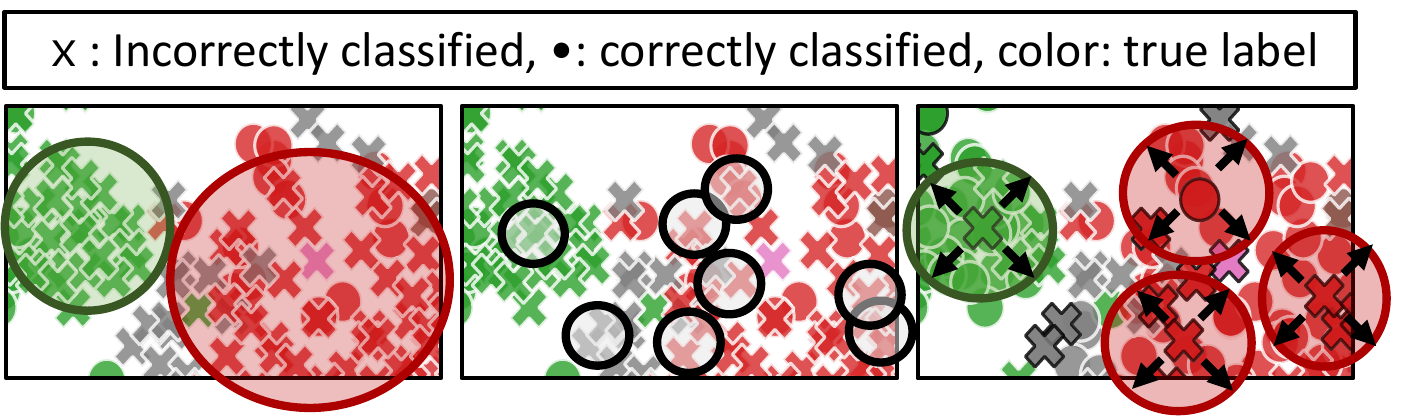}
    \text{(a) ~~~~~~~~~~~~~~~~~~~ (b) ~~~~~~~~~~~~~~~~~~~ (c)}
    \vspace{-0.2cm}
    \caption{
    \textbf{Why sparse labels can help VLM test-time adaptation.}
    \textbf{(a)} Under shift, a conventional VLM-TTA method can still make persistent errors on hard samples.
    \textbf{(b)} \alg\ selectively queries labels for a small number of uncertain samples and stores them as labeled anchors.
    \textbf{(c)} These anchors are then used to improve future predictions on related uncertain samples through retrieval-augmented correction.
    }
    \label{fig:problem}
    \vspace{-0.4cm}
\end{wrapfigure}
Vision-language models (VLMs)~\citep{radford2021learning,bai2023qwen,li2022blip,li2023blip,liu2023visual,bai2025qwen2} such as CLIP provide strong zero-shot recognition by aligning images and text in a shared embedding space, but their predictions can still degrade under deployment-time distribution shift. Test-time adaptation (TTA)~\citep{feng2023difftpt,shu2022tpt,yoon2024ctpt,karmanov2024efficient} addresses this problem without source data, and most VLM-TTA methods do so in a fully unsupervised way: they adapt from pseudo-labels, confidence scores, entropy, or caches produced by the same model being adapted. This label-free protocol is attractive, but it is also structurally limited on hard shifted samples. When the model is uncertain or miscalibrated, its self-generated signal can reinforce rather than repair an incorrect decision (See Figure~\ref{fig:problem}).

This paper studies a complementary regime between two extremes. Fully unsupervised VLM-TTA assumes no external supervision at deployment, while supervised online adaptation assumes labels can be used to update the model. Many practical systems lie between these extremes: a small amount of expert feedback may be available, but labels arrive after a prediction deadline and the deployed VLM must remain frozen. 
Existing active TTA methods such as simATTA~\citep{gui2024active} query labels for conventional classifiers and incorporate them through online parameter updates. This design typically operates at the batch/buffer level: predictions are first made for an incoming test batch, informative samples are selected for annotation, and the acquired labels are then used to update model parameters for subsequent batches. In contrast, our setting keeps the VLM frozen and treats queried labels as delayed, reusable evidence: once returned, they enter memory directly rather than being mediated by gradient-based adaptation.

\begin{figure}[t]
    \centering
    \includegraphics[width=0.9\linewidth]{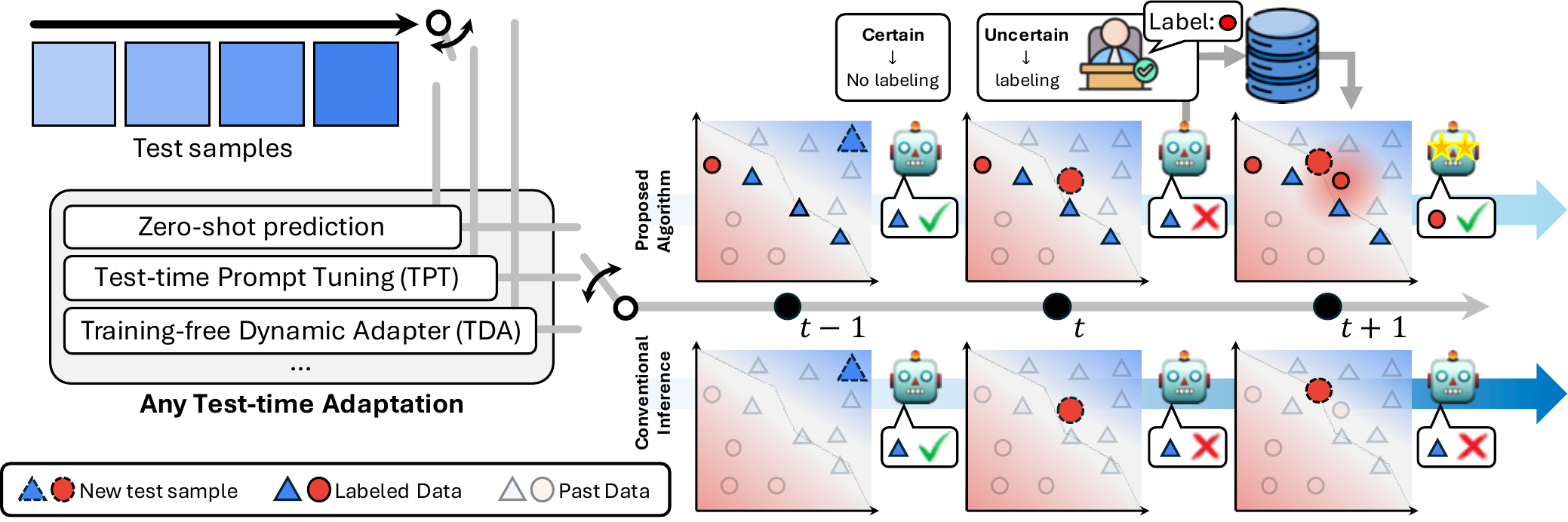}
    \caption{\textbf{Active test-time adaptation for vision-language model (Active VLM-TTA).} The model relies on a base VLM-TTA method for confident samples, while selectively querying labels for uncertain ones under a budget. Queried labels cannot revise the current prediction; instead, they are stored in labeled memory and used to improve future uncertain predictions. \alg\ instantiates this idea in a training-free and plug-and-play manner.}
    \vspace{-0.7cm}
    \label{fig:intro}
\end{figure}

We formalize this regime as \emph{Active Test-time Adaptation for VLMs} (\textbf{Active VLM-TTA}). At time $t$, the model must first predict on the current sample $x_t$; it may then query the label under a fixed budget, but that label can only be used for samples arriving after time $t$ (Figure~\ref{fig:intro}). Thus, the role of human supervision is deliberately constrained: labels are costed, sparse, and future-only. This formulation is not a replacement for label-free TTA, nor does it assume the convenience of a pre-existing support set. Instead, it asks how a pretrained VLM can convert delayed deployment-time labels into reusable evidence. VLMs are a natural fit for this question because their embedding spaces make semantically related examples retrievable across time, allowing a queried sample to serve as a labeled anchor for later uncertain samples.

To instantiate this setting, we propose \underline{\textbf{ACT}}ive VLM-TTA for \underline{\textbf{O}}nline \underline{\textbf{R}}etrieval (\alg), a training-free plug-in module for Active VLM-TTA. For each uncertain sample, \alg\ makes a three-way routing decision: use the base TTA prediction, query a labeler for future use, or correct the prediction from previously queried labels. The correction rule is intentionally conservative. In its main correction rule, a retrieved label is used only when the two nearest labeled anchors agree and are sufficiently close to the test sample in the VLM embedding space. This consistency-and-reliability check is a simple safeguard against the main failure mode of memory-based adaptation: propagating a sparse label beyond the neighborhood where it is trustworthy.

Empirically, \alg\ provides a strong first baseline for Active VLM-TTA. Across 10 cross-domain and 5 OOD benchmarks, it consistently improves CLIP zero-shot prediction and strong VLM-TTA methods on both ResNet and ViT backbones; with ViT-B/16, for example, it improves zero-shot CLIP ($61.94\% \rightarrow 70.67\%$) and TDA ($68.14\% \rightarrow 72.66\%$) on the cross-domain benchmark. Since labels are an explicit resource in our setup, we also compare against supervised prompt learning under the same queried labeled set: training-free \alg outperforms Online-CoOp and Replay-CoOp by 2.50 and 2.45 percentage points, respectively, across 15 datasets, while adding less than 1 ms of inference overhead. Our contributions are:

\begin{itemize}[leftmargin=*, itemsep=1pt, topsep=1pt]
    \item We introduce \textbf{Active VLM-TTA}, a streaming evaluation protocol for pretrained VLMs in which queried labels are budgeted and future-only, separating delayed human feedback from both label-free TTA and supervised online learning.
    \item We propose \alg, a training-free plug-in framework that converts sparse queried labels into reusable memory through three-way routing and a consistency-and-reliability correction rule.
    \item We provide the first broad empirical study of this regime, including label-matched supervised adaptation, budget, latency, hyperparameter, and VLM-family analyses.
\end{itemize}

%% file: revised_sections/02_Pre.tex
\section{Preliminary}
\label{sec:pre}

\subsection{Test-Time Adaptation with VLMs (VLM-TTA)}
\label{sec:tta}

CLIP~\citep{radford2021learning} consists of an image encoder $f_{\mathrm{img}}$ and a text encoder $f_{\mathrm{txt}}$, and performs classification by matching an image to class-specific text prompts. Given an input image $x_t$ and a class set $\mathcal{C}$, the similarity score for class $c \in \mathcal{C}$ is defined as
\begin{equation}
\label{eq:cos_sim}
    s_c = \mathrm{cos}\!\left(f_{\mathrm{img}}(x_t),\, f_{\mathrm{txt}}(t_c)\right),
\end{equation}
where $t_c$ denotes the text prompt for class $c$, and $\mathrm{cos}(\cdot,\cdot)$ is the cosine similarity. The corresponding prediction probability is
\begin{equation}
    z_t^c = \frac{\exp(s_c / \tau)}{\sum_{i=1}^{|\mathcal{C}|} \exp(s_i / \tau)},
\end{equation}
where $\tau$ is a temperature parameter. 

In the test-time adaptation (TTA) setting, we write the prediction of a VLM-TTA method in a unified form as
\begin{equation}
\label{eq:clip_prob}
    p(x_t) = \mathcal{T}_{\xi}(x_t; f_{\mathrm{img}}, f_{\mathrm{txt}}),
\end{equation}
where $\mathcal{T}_{\xi}(\cdot)$ denotes a TTA procedure built on the pretrained encoders $f_{\mathrm{img}}$ and $f_{\mathrm{txt}}$, together with any method-specific context variables $\xi$ such as learnable prompts (See Appendix~\ref{app:tta} for examples).

A common limitation of existing VLM-TTA methods is that their adaptation signals are derived from the model itself, such as pseudo-labels, confidence scores, or entropy. When the pretrained model is poorly calibrated under distribution shift, these self-generated signals can reinforce incorrect predictions rather than correct them.

\subsection{Problem Formulation: Active VLM-TTA}
\label{sec:prob}

We study a deployment setting in which the fully unsupervised assumption is relaxed, but only in a controlled way. The model may occasionally query ground-truth labels for incoming test samples under a limited budget. Crucially, the queried label is not available for revising the current prediction: the model must predict first, and any queried supervision can be used only for future samples. The label budget is therefore an explicit resource, and the queried labels are delayed evidence rather than free corrections. We formulate this setting as \textbf{Active VLM-TTA}, a streaming test-time adaptation protocol for frozen vision-language models with \emph{future-only supervision}.

\begin{definition}[Active VLM-TTA]
\label{def:active_vlm_tta}
For each test sample $x_t \in \mathfrak{D}_{\mathrm{test}}$ arriving at time $t$, a pretrained VLM must first output a prediction for $x_t$. It may then optionally query the ground-truth label $y_t$ under a labeling budget $\mathfrak{B}$. If $x_t$ is queried, its label is returned after a latency $\Delta_t \ge 1$. 
Until then, the queried sample is stored in a pending set rather than in the usable labeled memory. 
At time $t$, the labeled memory is updated only with labels whose latency has expired:
\[
D_l(t+1)=D_l(t)\cup 
\{(x_s,y_s): s+\Delta_s \le t+1,\; x_s \text{ was queried}\}.
\]
Thus, the queried label for $x_t$ cannot affect the prediction on $x_t$ and becomes available only for samples arriving after its return time. 
\end{definition}

Active VLM-TTA differs from standard online active learning in that the objective is not to improve a continually updated classifier through label-efficient training, but to improve online test-time inference while preserving a frozen VLM. It also differs from prior active test-time adaptation for conventional classifiers~\citep{gui2024active}, which relies on online parameter updates. This makes the setting complementary to label-free VLM-TTA: the question is not whether labels help, but how a small, explicitly costed set of queried labels should be reused under streaming constraints.

%% file: revised_sections/03_methodology.tex
\section{\alg: \texorpdfstring{\underline{ACT}}iive VLM-TTA for \texorpdfstring{\underline{O}}nnline \texorpdfstring{\underline{R}}eetrieval}
\label{sec:method}

\noindent The key challenges of Active VLM-TTA are (1) deciding which uncertain samples are worth spending the label budget on and (2) reusing the resulting labels without over-propagating them to unrelated future samples. \alg{} is designed as a conservative training-free baseline for this protocol: it changes neither the VLM parameters nor the base TTA method, and intervenes only when delayed labeled evidence is sufficiently local and consistent. Its process is summarized as follows (see Appendix~\ref{app:alg} for the full design).

\begin{enumerate}[leftmargin=* , itemsep=0pt, topsep=2pt]
    \item Filter uncertain samples likely to be misclassified. $\triangleright$~Section~\ref{subsec:uncert_filter}
    \item Among uncertain samples, build $\mathcal{D}_l(t)$ until it reaches size $N$.
    \begin{enumerate}[label=(\alph*), leftmargin=1.5em, itemsep=0pt, topsep=1pt]
        \item If consistency and reliability are high, apply retrieval-augmented correction. $\triangleright$~Section~\ref{subsec:retr_pred}
        \item Otherwise, query the label and add the sample to $\mathcal{D}_l(t)$ for future use once the annotation is returned. In case of budget exhaustion, apply retrieval-augmented correction under related criteria $\triangleright$~Section~\ref{subsec:saturated}
    \end{enumerate}
\end{enumerate}

\subsection{Uncertainty-based Filtering}
\label{subsec:uncert_filter}
We first choose the samples that the model is most likely to misclassify.
For the test sample \(x_t\), we first compute the base TTA prediction probability \(q_t = T_\xi(x_t; f_{\rm img}, f_{\rm txt})\). If \(T_\xi\) outputs logits, we apply softmax normalization to obtain \(q_t\). We then calculate entropy as
\begin{equation}
    \label{eq:entropy}
    \mathcal{H}(q_t)=-\sum_{c\in C} q_{t,c}\log q_{t,c}.
\end{equation}
Since the entropy is a measure of \textit{how uncertain a model is in its prediction for a given sample}, a sample with a high-entropy value is more likely to be predicted incorrectly by the model~\citep{holub2008entropy}. 
Therefore, we conduct retrieval-augmented correction only for high-entropy samples.

\subsection{Retrieval-augmented Correction}
\label{subsec:retr_pred}

We employ a retrieval-augmented correction strategy that leverages $\mathfrak{D}_l(t)$ to correct predictions made by the base TTA method. However, a sparse queried label is useful only within the part of the embedding space where it remains trustworthy; applying nearest-neighbor correction to every uncertain sample can easily introduce new errors. We therefore use two criteria before accepting a memory-based correction: \textbf{consistency}, which checks whether the labeled samples nearest to a test sample agree in label space, and \textbf{reliability}, which measures how close those labeled samples are to the test sample. Intuitively, if the nearest labeled samples share a label and are close to the current sample, their shared label is more plausible as delayed evidence for the current prediction. For simplicity, we refer to the first and second nearest labeled samples as a \emph{Referrer} $\mathrm{R} = (x_t^R, y_t^R)$ and a \emph{Verifier} $\mathrm{V} = (x_t^V, y_t^V)$, respectively.



\begin{figure}[t]
    \centering 
    
    \begin{subfigure}[b]{0.27\linewidth}
        \centering
        \includegraphics[width=\textwidth]{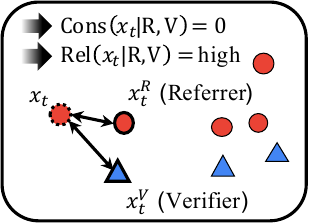}
        \caption{} 
        \label{fig:ref_ver_a}
    \end{subfigure}%
    \hspace{0.3cm}%
    \begin{subfigure}[b]{0.27\linewidth}
        \centering
        \includegraphics[width=\textwidth]{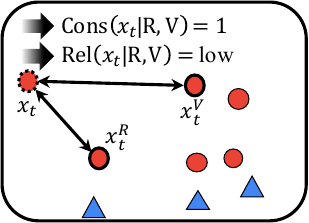}
        \caption{} 
        \label{fig:ref_ver_b}
    \end{subfigure}%
    \hspace{0.3cm}%
    \begin{subfigure}[b]{0.27\linewidth}
        \centering
        \includegraphics[width=\textwidth]{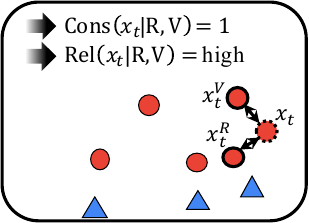}
        \caption{} 
        \label{fig:ref_ver_c}
    \end{subfigure}
    
    \vspace{-0.15cm}
    \caption{\textbf{Conceptual illustration of retrieval-augmented correction.} \textbf{(a)} Low consistency. \textbf{(b)} High consistency, but low reliability. \textbf{(c)} High consistency, and  high reliability. \alg operates under the philosophy of reflecting predictions based on proximity to a given test sample, provided the labels between neighboring samples are consistent.}
    \label{fig:ref_ver}
    \vspace{-0.3cm}
\end{figure}

Figure~\ref{fig:ref_ver} depicts three common cases for the Referrer-Verifier interactions. In Figure~\ref{fig:ref_ver_a}, the Referrer and Verifier disagree on the label, so no retrieval-augmented correction is applied. In Figure~\ref{fig:ref_ver_b}, the labels match, but the Verifier is far from the test sample, making its endorsement of the Referrer's label less trustworthy. In contrast, Figure~\ref{fig:ref_ver_c} shows both high consistency (same label) and high reliability (all close), thus allowing retrieval-augmented correction to proceed confidently. Based on these common cases, we now provide the detailed mathematical formulation as follows.


\myparagraph{Consistency.} To ensure that the candidate label from the Referrer is reliable, we define the \emph{consistency} metric to check whether the Verifier's label agrees with the Referrer's. Formally, it is defined as
\begin{equation}
\label{eq:cons} 
\mathrm{Cons}(x_t| \mathrm{R}, \mathrm{V}) = \mathbbm{1}\{ y_t^{R} = y_t^{V} \}, 
\end{equation} 
where $\mathbbm{1}\{\cdot\}$ is the indicator function. A value of 1 indicates full agreement between the Referrer and the Verifier, thus confirming the candidate label.
This check is motivated by a simple design principle: agreement between two nearby labeled anchors is less likely to reflect an isolated retrieval error than relying on a single nearest neighbor alone.

\myparagraph{Reliability.} 
To quantify the joint proximity of the Referrer and Verifier to a test sample $x_t$, we define reliability as the average cosine similarity between the test sample and its two nearest labeled neighbors. Formally, this metric is given by
\begin{equation} 
\label{eq:rel} 
\begin{split}
\mathrm{Rel}(x_t | \mathrm{R}, \mathrm{V}) = \frac{1}{2} \Big( \mathrm{cos}\big(f_{\mathrm{img}}&(x_t),f_{\mathrm{img}}(x_t^{R})\big) + 
\mathrm{cos}\big(f_{\mathrm{img}}(x_t), f_{\mathrm{img}}(x_t^{V})\big) \Big). 
\end{split}
\end{equation} 
The reliability threshold further restricts this agreement to local neighborhoods in the VLM embedding space, reducing the chance that a queried label is propagated too far from the region where it is informative.

\paragraph{Overall correction criterion.}
The condition for retrieval-augmented correction is formulated as
\begin{equation}
C(x_t \mid R,V) = \mathrm{Cons}(x_t \mid R,V)\cdot \mathrm{Rel}(x_t \mid R,V).
\label{eq:criteria}
\end{equation}
This multiplicative formulation ensures that if the Verifier disagrees with the Referrer, i.e., $\mathrm{Cons}(x_t \mid R,V)=0$, no retrieval-based correction is applied. Taken together, the consistency check suppresses locally noisy retrievals, while the reliability threshold restricts correction to nearby labeled evidence. This motivates our conservative routing rule, which activates retrieval-based correction only when both conditions are satisfied.

\subsection{Correction after Label Budget Exhaustion}
\label{subsec:saturated}

Once the labeling budget $\mathfrak{B}$ is exhausted, no additional human feedback can be obtained. In this regime, \alg retains the consistency-and-reliability correction rule in Section~\ref{subsec:retr_pred} and applies a relaxed Top-$K$ plausibility check as a fallback when the original criterion is not satisfied.
Let $q_t$ be the class-probability vector predicted by the base TTA model for sample $x_t$, and let $y_t^R$ denote the label suggested by the Referrer. We accept $y_t^R$ only if it belongs to $\mathrm{Top}\text{-}K(q_t)$, the set of classes with the $K$ largest probabilities under $q_t$:

\begin{equation}
\label{eq:budget}
\hat{p}_t =
\begin{cases}
\mathrm{one\_hot}(y_t^R), 
& \text{if }  y_t^R \in \mathrm{Top}\text{-}K(q_t),\\
q_t, & \text{otherwise}.
\end{cases}
\end{equation}

This fallback is a label-space plausibility check after the annotation budget is exhausted, not the
full spatial reliability test used before budget exhaustion. It avoids unrestricted nearest-neighbor
correction by accepting the retrieved label only when it appears among the \(K\) highest-confidence
labels of the base TTA prediction \(q_t\). We set \(K=5\) across all datasets, following the standard
Top-5 criterion in large-scale image classification~\citep{deng2009imagenet}.

%% file: revised_sections/04_exp.tex
\setlength{\tabcolsep}{8pt}
\begin{table}[t]
  \vspace{-0.4cm}
  \centering
  \scriptsize
  \begin{tabular}{lccccccc}
    \toprule
     \textbf{Method}                        &  \multicolumn{1}{c}{IN} & \multicolumn{1}{c}{IN-A} & \multicolumn{1}{c}{IN-V2} & \multicolumn{1}{c}{IN-R} &\multicolumn{1}{c}{IN-S}  & \multicolumn{1}{c}{Avg (All5)} & \multicolumn{1}{c}{Avg(OOD)}\\
    \hline \hline
    \multicolumn{8}{c}{\textbf{ResNet-50}}\\ \hline
    Zero-shot                               &    56.47    &    26.96    &    50.83    &    55.87    &    31.50    &  44.33      &    41.29   \\ 
    \rowcolor[gray]{.9} $~~ + \mathrm{\alg{}}$                  &    57.66{\tiny $\pm$0.06}    &    28.11{\tiny $\pm$0.19}    &    50.76{\tiny $\pm$0.07}    &    56.93{\tiny $\pm$0.07}    &    36.94{\tiny $\pm$0.04}    &  46.08      &    43.18   \\\cmidrule(lr){1-8}
    TPT                                     &    60.68{\tiny $\pm$0.01}    &    25.12{\tiny $\pm$0.07}    &    54.33{\tiny $\pm$0.23}    &    59.11{\tiny $\pm$0.04}    &    35.29{\tiny $\pm$0.04}    &  46.91      &    43.46   \\ 
    \rowcolor[gray]{.9} $~~ + \mathrm{\alg{}}$                  &    61.20{\tiny $\pm$0.10}    &    26.28{\tiny $\pm$0.05}    &    54.34{\tiny $\pm$0.02}    &    59.65{\tiny $\pm$0.18}    &    39.67{\tiny $\pm$0.18}    &  48.23      &    44.98  \\\cmidrule(lr){1-8}
    C-TPT                                   &    60.42{\tiny $\pm$0.06}    &    23.25{\tiny $\pm$0.06}    &    54.13{\tiny $\pm$0.13}    &    57.72{\tiny $\pm$0.01}    &    34.76{\tiny $\pm$0.01}    &  46.05      &    42.46   \\ 
    \rowcolor[gray]{.9} $~~ + \mathrm{\alg{}}$                  &    61.10{\tiny $\pm$0.11}    &    24.76{\tiny $\pm$0.30}    &    53.88{\tiny $\pm$0.03}    &    58.66{\tiny $\pm$0.12}    &    39.62{\tiny $\pm$0.02}    &  47.60      &    44.23   \\\cmidrule(lr){1-8}
    TDA$^\dagger$                           &    57.78{\tiny $\pm$0.01}    &    27.84{\tiny $\pm$0.01}    &    51.10{\tiny $\pm$0.03}    &    57.01{\tiny $\pm$0.01}    &    33.85{\tiny $\pm$0.11}    &  45.52      &    42.45   \\ 
    \rowcolor[gray]{.9} $~~ + \mathrm{\alg{}}$                  &    58.64{\tiny $\pm$0.01}    &    28.73{\tiny $\pm$0.28}    &    51.31{\tiny $\pm$0.09}    &    57.84{\tiny $\pm$0.06}    &    39.00{\tiny $\pm$0.01}    &  47.10      &    44.22   \\\cmidrule(lr){1-8}
    TDA                                     &    61.72{\tiny $\pm$0.04}    &    30.59{\tiny $\pm$0.21}    &    \textbf{55.10}{\tiny $\pm$0.26}    &    62.76{\tiny $\pm$0.04}    &    38.05{\tiny $\pm$0.04}    &  49.64      &    46.62   \\ 
    \rowcolor[gray]{.9} $~~ + \mathrm{\alg{}}$                  &    \textbf{62.06}{\tiny $\pm$0.04}    &    \textbf{31.00}{\tiny $\pm$0.15}    &    54.99{\tiny $\pm$0.29}    &    \textbf{63.25}{\tiny $\pm$0.12}    &    \textbf{42.73}{\tiny $\pm$0.10}    &  \textbf{50.80}      &    \textbf{47.99}   \\

    \hline \hline
    
    \multicolumn{8}{c}{\textbf{ViT-B/16}}\\ \hline
    Zero-shot                               &    63.52    &    54.65    &    59.80    &    75.36    &    44.05    &    59.48    &    58.47   \\ 
    \rowcolor[gray]{.9} $~~ + \mathrm{\alg{}}$                  &    65.90{\tiny $\pm$1.03}    &    55.85{\tiny $\pm$0.74}    &    59.69{\tiny $\pm$0.14}    &    76.33{\tiny $\pm$0.02}    &    50.29{\tiny $\pm$0.13}    &    61.61    &    60.54   \\\cmidrule(lr){1-8}
    TPT                                     &    68.76{\tiny $\pm$0.01}    &    52.99{\tiny $\pm$0.06}    &    63.20{\tiny $\pm$0.04}    &    76.97{\tiny $\pm$0.10}    &    47.86{\tiny $\pm$0.02}    &    61.95    &    60.25   \\ 
    \rowcolor[gray]{.9} $~~ + \mathrm{\alg{}}$                  &    69.27{\tiny $\pm$0.12}    &    54.89{\tiny $\pm$0.05}    &    61.94{\tiny $\pm$1.49}    &    77.18{\tiny $\pm$0.08}    &    51.30{\tiny $\pm$0.94}    &    62.91    &    61.32  \\\cmidrule(lr){1-8}
    C-TPT                                   &    68.31{\tiny $\pm$0.01}    &    50.71{\tiny $\pm$0.04}    &    62.50{\tiny $\pm$0.04}    &    75.68{\tiny $\pm$0.08}    &    47.46{\tiny $\pm$0.06}    &    60.93    &    59.08   \\ 
    \rowcolor[gray]{.9} $~~ + \mathrm{\alg{}}$                  &    69.09{\tiny $\pm$0.01}    &    53.09{\tiny $\pm$0.17}    &    62.14{\tiny $\pm$0.03}    &    76.39{\tiny $\pm$0.13}    &    52.13{\tiny $\pm$0.01}    &    62.57    &    60.94   \\\cmidrule(lr){1-8}
    TDA$^\dagger$                           &    65.81{\tiny $\pm$0.81}    &    55.42{\tiny $\pm$0.06}    &    60.06{\tiny $\pm$0.13}    &    75.99{\tiny $\pm$0.13}    &    46.01{\tiny $\pm$0.08}    &    60.66    &    59.37   \\ 
    \rowcolor[gray]{.9} $~~ + \mathrm{\alg{}}$                  &    67.48{\tiny $\pm$0.03}    &    56.49{\tiny $\pm$0.20}    &    60.09{\tiny $\pm$0.12}    &    76.70{\tiny $\pm$0.08}    &    51.23{\tiny $\pm$0.02}    &    62.40    &    61.13   \\\cmidrule(lr){1-8}
    TDA                                     &    70.08{\tiny $\pm$0.88}    &    60.13{\tiny $\pm$0.33}    &    64.41{\tiny $\pm$0.16}    &    80.52{\tiny $\pm$0.06}    &    50.96{\tiny $\pm$0.06}    &    65.22    &    64.01   \\ 
    \rowcolor[gray]{.9} $~~ + \mathrm{\alg{}}$                  &    \textbf{70.88}{\tiny $\pm$0.11}    &    \textbf{60.73}{\tiny $\pm$0.60}    &    \textbf{64.54}{\tiny $\pm$0.10}    &    \textbf{80.83}{\tiny $\pm$0.11}    &    \textbf{55.01}{\tiny $\pm$0.07}    &    \textbf{66.40}    &    \textbf{65.28}   \\

    \bottomrule
    \end{tabular}
  \caption{\textbf{Results on the out-of-distribution benchmark.} All the compared methods are built upon CLIP-ResNet-50 or CLIP-ViT-B/16 baselines. The two average metrics \emph{All 5} and \emph{OOD} are calculated by taking the mean accuracy across all five datasets and four OOD datasets excluding ImageNet (IN). Note that the \textbf{bold} type represents the best performance overall. $\dagger$ indicates that the method uses the \emph{unified} prompt template (\ie a single ``a photo of a \{class\}.'') that is shared with TPT and C-TPT, rather than TDA's original per-dataset hand-crafted templates.
  } 
  \vspace{-0.7cm}
  \label{tab:ood_main}
\end{table}

\section{Experiment}
\label{sec:exp}

We provide details of the experimental setup, including datasets, baselines, and implementation details, in Appendix~\ref{app:exp} due to space constraints. This section focuses on CLIP-based main results and ablations; Appendix~\ref{app:further_analyses} includes results on BLIP-2~\citep{li2023blip} and SigLIP~\citep{zhai2023sigmoid}, as well as an ablation over uncertainty metrics. Since Active VLM-TTA explicitly allows a limited amount of delayed human supervision, we report the labeling budget as part of the protocol rather than treating labels as free. 
Unless otherwise stated, all ablations use CLIP-ResNet-50 with a fixed validation-free setting: \(\tau_u=0.2\), \(\tau_r=0.85\), \(N=5\), \(\Delta_t=1\), and a 10\% labeling budget. Other backbones are explicitly indicated where used.

\subsection{Main Results}
\myparagraph{Consistent gains under costed delayed labels.}
\alg{} can be integrated into any VLM-TTA pipeline, yielding consistent improvements under the same future-only labeling protocol. As shown in Table~\ref{tab:ood_main}, adding \alg{} to a baseline VLM-TTA method improves top-1 accuracy on the OOD benchmark, especially on ImageNet-S (50.96\%$\rightarrow$55.01\%), which raises the Avg(OOD) score (64.01\%$\rightarrow$65.28\%). On the cross-domain benchmark in Table~\ref{tab:cd_main}, \alg{} also delivers consistent gains, with a particularly large improvement on Flowers102 (71.62\%$\rightarrow$82.28\%). These results should be interpreted as the value of explicitly budgeted, delayed label reuse in Active VLM-TTA, not as a label-free replacement for unsupervised TTA.

\setlength{\tabcolsep}{1pt}
\begin{table*}[t]
  \centering
  \scriptsize
  \begin{tabular}{lccccccccccc}
    \toprule
     \textbf{Method}                        &  \multicolumn{1}{c}{Aircraft} & \multicolumn{1}{c}{Caltech101} & \multicolumn{1}{c}{Cars} & \multicolumn{1}{c}{DTD} &\multicolumn{1}{c}{EuroSAT}  & \multicolumn{1}{c}{Flowers102} & \multicolumn{1}{c}{Food101} & \multicolumn{1}{c}{Pets} & \multicolumn{1}{c}{SUN397} & \multicolumn{1}{c}{UCF101} & \multicolumn{1}{c}{Average} \\ 
     \hline \hline
      \multicolumn{12}{c}{\textbf{ResNet50}} \\ \hline

    Zero-shot                                   &    15.30    &    80.61    &    56.27    &    37.53    &    25.89    &    56.44    &    73.45    &    80.46    &    58.16    &    59.16    &    54.33  \\ 
    \rowcolor[gray]{.9} $~~ + \mathrm{\alg{}}$                      &    18.16{\tiny $\pm$0.25}    &    84.61{\tiny $\pm$0.26}    &    57.54{\tiny $\pm$0.42}    &    43.35{\tiny $\pm$1.05}    &    \textbf{74.16}{\tiny $\pm$1.12}    &    72.55{\tiny $\pm$0.06}    &    73.73{\tiny $\pm$0.08}    &    81.71{\tiny $\pm$0.11}    &    59.41{\tiny $\pm$0.08}    &    70.15{\tiny $\pm$0.62}    &    63.53\\\cmidrule(lr){1-12}
    TPT                                         &    16.02{\tiny $\pm$0.47}    &    85.36{\tiny $\pm$0.40}    &    57.95{\tiny $\pm$0.16}    &    39.69{\tiny $\pm$0.37}    &    28.04{\tiny $\pm$0.03}    &    60.22{\tiny $\pm$0.40}    &    73.63{\tiny $\pm$0.02}    &    78.39{\tiny $\pm$0.12}    &    59.86{\tiny $\pm$0.12}    &    60.77{\tiny $\pm$0.34}    &    55.99  \\ 
    \rowcolor[gray]{.9} $~~ + \mathrm{\alg{}}$                      &    18.60{\tiny $\pm$0.25}    &    87.73{\tiny $\pm$0.83}    &    \textbf{60.31}{\tiny $\pm$0.52}    &    45.16{\tiny $\pm$0.50}    &    62.16{\tiny $\pm$0.69}    &    70.61{\tiny $\pm$0.40}    &    73.89{\tiny $\pm$0.04}    &    80.86{\tiny $\pm$0.02}    &    61.45{\tiny $\pm$0.13}    &    69.20{\tiny $\pm$0.52}    &    62.99  \\\cmidrule(lr){1-12}
    C-TPT                                       &    13.70{\tiny $\pm$0.15}    &    85.82{\tiny $\pm$0.20}    &    56.51{\tiny $\pm$0.06}    &    40.58{\tiny $\pm$0.29}    &    23.13{\tiny $\pm$0.45}    &    61.29{\tiny $\pm$0.21}    &    73.13{\tiny $\pm$0.04}    &    80.04{\tiny $\pm$0.13}    &    59.54{\tiny $\pm$0.04}    &    59.31{\tiny $\pm$0.02}    &    55.30  \\
    \rowcolor[gray]{.9} $~~ + \mathrm{\alg{}}$                      &    16.65{\tiny $\pm$0.51}    &    87.59{\tiny $\pm$0.69}    &    59.35{\tiny $\pm$0.11}    &    45.89{\tiny $\pm$0.96}    &    60.76{\tiny $\pm$0.78}    &    73.02{\tiny $\pm$1.23}    &    73.64{\tiny $\pm$0.03}    &    81.85{\tiny $\pm$0.38}    &    61.28{\tiny $\pm$0.45}    &    69.83{\tiny $\pm$0.25}    &    62.98\\\cmidrule(lr){1-12}
    TDA$^\dagger$                              &    15.93{\tiny $\pm$0.81}    &    85.13{\tiny $\pm$0.03}    &    57.17{\tiny $\pm$0.08}    &    39.34{\tiny $\pm$0.29}    &    36.71{\tiny $\pm$1.01}    &    59.24{\tiny $\pm$0.11}    &    74.62{\tiny $\pm$0.04}    &    80.69{\tiny $\pm$0.10}    &    59.68{\tiny $\pm$0.11}    &    60.75{\tiny $\pm$0.26}    &    56.92 \\ 
    \rowcolor[gray]{.9} $~~ + \mathrm{\alg{}}$                      &    17.75{\tiny $\pm$0.74}    &    85.52{\tiny $\pm$0.23}    &    58.48{\tiny $\pm$0.09}    &    44.18{\tiny $\pm$0.79}    &    70.81{\tiny $\pm$1.08}    &    71.64{\tiny $\pm$0.66}    &    74.63{\tiny $\pm$0.19}    &    81.84{\tiny $\pm$0.06}    &    60.40{\tiny $\pm$0.04}    &    68.86{\tiny $\pm$0.98}    &    63.41 \\\cmidrule(lr){1-12}
    TDA                                         &    16.79{\tiny $\pm$0.23}    &    89.36{\tiny $\pm$0.43}    &    57.36{\tiny $\pm$0.17}    &    43.91{\tiny $\pm$0.08}    &    41.66{\tiny $\pm$0.61}    &    68.17{\tiny $\pm$0.28}    &    \textbf{77.78}{\tiny $\pm$0.11}    &    86.29{\tiny $\pm$0.27}    &    62.48{\tiny $\pm$0.01}    &    64.03{\tiny $\pm$0.15}    &    60.78\\ 
    \rowcolor[gray]{.9} $~~ + \mathrm{\alg{}}$                      &    \textbf{19.01}{\tiny $\pm$0.02}    &    \textbf{89.57}{\tiny $\pm$0.01}    &    58.67{\tiny $\pm$0.33}    &    \textbf{48.17}{\tiny $\pm$0.08}    &    71.08{\tiny $\pm$1.29}    &    \textbf{77.19}{\tiny $\pm$0.63}    &    77.28{\tiny $\pm$0.20}    &    \textbf{86.62}{\tiny $\pm$0.50}    &    \textbf{62.81}{\tiny $\pm$0.11}    &    \textbf{71.27}{\tiny $\pm$1.12}    &    \textbf{66.16} \\

    \hline \hline 
     \multicolumn{12}{c}{\textbf{ViT-B/16}}\\ \hline
    Zero-shot                                   &    22.59    &    85.76    &    65.61    &    40.60    &    44.25    &    64.07    &    82.68    &    83.81    &    63.68    &    66.30    &    61.94  \\ 
    \rowcolor[gray]{.9} $~~ + \mathrm{\alg{}}$                      &    27.33{\tiny $\pm$0.71}    &    90.31{\tiny $\pm$0.12}    &    67.77{\tiny $\pm$0.25}    &    47.94{\tiny $\pm$0.42}    &    \textbf{80.63}{\tiny $\pm$0.54}    &    80.53{\tiny $\pm$0.20}    &    83.57{\tiny $\pm$0.06}    &    87.10{\tiny $\pm$0.13}    &    65.30{\tiny $\pm$0.03}    &    76.28{\tiny $\pm$0.54}    &    70.67\\\cmidrule(lr){1-12}
    TPT                                         &    23.43{\tiny $\pm$0.08}    &    93.21{\tiny $\pm$1.07}    &    66.52{\tiny $\pm$0.20}    &    46.33{\tiny $\pm$1.10}    &    42.69{\tiny $\pm$0.16}    &    68.25{\tiny $\pm$0.93}    &    83.94{\tiny $\pm$1.18}    &    84.42{\tiny $\pm$0.98}    &    65.19{\tiny $\pm$0.47}    &    67.62{\tiny $\pm$0.41}    &    64.16  \\ 
    \rowcolor[gray]{.9} $~~ + \mathrm{\alg{}}$                      &    25.56{\tiny $\pm$0.47}    &    93.37{\tiny $\pm$0.08}    &    67.97{\tiny $\pm$0.04}    &    49.53{\tiny $\pm$0.01}    &    73.03{\tiny $\pm$0.33}    &    74.63{\tiny $\pm$0.80}    &    83.54{\tiny $\pm$0.04}    &    85.96{\tiny $\pm$0.16}    &    66.49{\tiny $\pm$0.06}    &    72.67{\tiny $\pm$0.23}    &    69.27\\\cmidrule(lr){1-12}
    C-TPT                                       &    24.09{\tiny $\pm$0.25}    &    92.98{\tiny $\pm$0.69}    &    65.24{\tiny $\pm$0.78}    &    45.01{\tiny $\pm$1.21}    &    42.28{\tiny $\pm$0.10}    &    70.36{\tiny $\pm$0.75}    &    83.22{\tiny $\pm$0.11}    &    87.82{\tiny $\pm$0.72}    &    64.07{\tiny $\pm$0.59}    &    65.58{\tiny $\pm$0.79}    &    64.06  \\
    \rowcolor[gray]{.9} $~~ + \mathrm{\alg{}}$                      &    27.78{\tiny $\pm$0.47}    &    93.27{\tiny $\pm$0.18}    &    66.88{\tiny $\pm$0.37}    &    50.59{\tiny $\pm$0.17}    &    74.72{\tiny $\pm$0.79}    &    77.49{\tiny $\pm$0.43}    &    83.73{\tiny $\pm$0.05}    &    87.70{\tiny $\pm$0.02}    &    66.22{\tiny $\pm$0.05}    &    74.65{\tiny $\pm$0.64}    &    70.30\\\cmidrule(lr){1-12}
    TDA$^\dagger$                               &    23.48{\tiny $\pm$0.28}    &    88.54{\tiny $\pm$0.08}    &    66.59{\tiny $\pm$0.11}    &    42.97{\tiny $\pm$0.67}    &    54.82{\tiny $\pm$0.77}    &    65.27{\tiny $\pm$0.49}    &    83.82{\tiny $\pm$0.08}    &    85.12{\tiny $\pm$0.27}    &    65.14{\tiny $\pm$0.01}    &    69.63{\tiny $\pm$0.15}    &    64.54 \\ 
    \rowcolor[gray]{.9} $~~ + \mathrm{\alg{}}$                      &    25.74{\tiny $\pm$0.21}    &    90.41{\tiny $\pm$0.43}    &    67.94{\tiny $\pm$0.04}    &    49.41{\tiny $\pm$0.34}    &    75.40{\tiny $\pm$0.54}    &    79.20{\tiny $\pm$0.49}    &    84.20{\tiny $\pm$0.06}    &    87.37{\tiny $\pm$0.10}    &    66.18{\tiny $\pm$0.01}    &    76.01{\tiny $\pm$0.62}    &    70.18\\\cmidrule(lr){1-12}
    TDA                                         &    25.29{\tiny $\pm$0.30}    &    \textbf{94.06}{\tiny $\pm$0.08}    &    66.39{\tiny $\pm$1.72}    &    45.48{\tiny $\pm$0.79}    &    63.95{\tiny $\pm$0.92}    &    71.62{\tiny $\pm$0.23}    &    \textbf{86.14}{\tiny $\pm$0.06}    &    89.85{\tiny $\pm$0.13}    &    67.70{\tiny $\pm$0.04}    &    70.94{\tiny $\pm$0.28}    &    68.14\\ 
    \rowcolor[gray]{.9} $~~ + \mathrm{\alg{}}$                      &    \textbf{27.84}{\tiny $\pm$0.08}    &    \textbf{94.06}{\tiny $\pm$0.03}    &    \textbf{68.76}{\tiny $\pm$0.13}    &    \textbf{51.63}{\tiny $\pm$0.88}    &    79.61{\tiny $\pm$1.40}    &    \textbf{82.28}{\tiny $\pm$0.54}    &    86.09{\tiny $\pm$0.06}    &    \textbf{90.75}{\tiny $\pm$0.18}    &    \textbf{68.20}{\tiny $\pm$0.11}    &    \textbf{77.38}{\tiny $\pm$1.01}    &    \textbf{72.66} \\

    \bottomrule
    \end{tabular}

    \vspace*{-0.2cm}
  \caption{\textbf{Results on the cross-domain benchmark.} The \emph{Average} is calculated by taking the mean accuracy across all ten datasets. Note that the \textbf{bold} type represents the best performance overall. $\dagger$ indicates that the method uses the \emph{unified} prompt template (\ie a single ``a photo of a \{class\}.'') that is shared with TPT and C-TPT, rather than TDA's original per-dataset hand-crafted templates.
  } 
  \vspace*{-0.3cm}
  \label{tab:cd_main}
\end{table*}

\myparagraph{Architecture-agnostic efficacy.} \alg{} is effective across different backbone architectures. On both the OOD and cross-domain benchmarks, integrating \alg{} consistently improves performance regardless of whether the base model is ResNet-based or Transformer-based. For the OOD benchmark in Table~\ref{tab:ood_main}, applying \alg{} to TDA with a ResNet-50 backbone improves the average accuracy across all five datasets (49.64\%$\rightarrow$50.80\%), while the ViT-B/16 backbone shows a comparable gain (65.22\%$\rightarrow$66.40\%). Likewise, on the cross-domain benchmark in Table~\ref{tab:cd_main}, the ResNet-50 model enjoys a substantial accuracy gain (60.78\%$\rightarrow$66.16\%), and the ViT-B/16 model achieves a similar boost (68.14\%$\rightarrow$72.66\%). These results indicate that \alg{} is a robust enhancement for both CNN- and Transformer-based VLM backbones.


\myparagraph{Streaming comparison with online CoOp-style prompt tuning under matched supervision.}
To verify that \alg's gains are not merely due to access to queried labels, we compare it with two supervised prompt-tuning baselines under the same Active VLM-TTA protocol. Online-CoOp performs at most one gradient update per queried sample, whereas Replay-CoOp reuses accumulated labeled samples for subsequent updates. Both baselines share the same queried labels, labeling budget, and future-only constraint as \alg, but rely on gradient-based prompt optimization rather than training-free retrieval memory. As shown in Table~\ref{tab:train}, \alg improves accuracy across 15 datasets by 2.50 and 2.45 percentage points (pp) over Online-CoOp and Replay-CoOp, respectively. These results indicate that \alg's gains cannot be explained solely by label access or repeated label reuse.

\myparagraph{Inference cost.}
For a fair comparison, we run all experiments on Intel Xeon Gold 6326 CPUs with a single NVIDIA RTX 4090. After a warm-up with 100 samples, we measure the time taken to process the next 100 samples, as reported in Table~\ref{tab:time_cost}. This table measures computational inference overhead; human labeling cost is accounted for separately through the explicit budget and latency analyses. Training-based TTA methods (\eg TPT, C-TPT) incur an additional cost of roughly 500\,ms per sample over train-free methods (\eg TDA). In contrast, \alg{} introduces only a negligible computational overhead of \emph{below 1\,ms} per sample.

\setlength{\tabcolsep}{3pt}
\begin{table*}[t]    
\begin{minipage}{0.63\textwidth}
    \centering
    \scriptsize
    \begin{tabular}{lcccc}
    \toprule
    \textbf{Method}
    & \textbf{Cross (10)}
    & \textbf{OOD (4)}
    & \textbf{OOD (5)}
    & \textbf{Avg. (15)} \\
    \midrule
    CLIP (Zero-shot)
    & 61.94 & 58.47 & 59.48 & 61.12 \\

    CLIP + Online-CoOp
    & 67.43 & 58.61 & 60.58 & 65.15 \\

    CLIP + Replay-CoOp
    & 69.03 & 56.24 & 57.54 & 65.20 \\

    \rowcolor[gray]{.9}
    CLIP + \alg{} (Train-free)
    & \textbf{70.67}
    & \textbf{60.54}
    & \textbf{61.61}
    & \textbf{67.65} \\
    \bottomrule
    \end{tabular}

    \captionof{table}{
    \textbf{Label-matched comparison with online CoOp baselines (ViT-B/16).}
    Cross and OOD denote average accuracy across 10 and 4/5 datasets, respectively. Online-CoOp performs one update per queried label, while Replay-CoOp reuses accumulated labels. All label-using methods share identical queried labels under future-only supervision  (See Appendix~\ref{app:online_prompt_tuning}).
    }
    
    \label{tab:train}
    \vspace{-0.4cm}
\end{minipage}
\hfill
\begin{minipage}{0.35\textwidth}
    \vspace{-0.8em}
    \centering
    \scriptsize
    \begin{tabular}{@{}ccccc@{}}
    
        \toprule
        \textbf{\alg{}}                        &       \multicolumn{1}{c}{Zero-shot}           &    \multicolumn{1}{c}{TPT}                &       \multicolumn{1}{c}{C-TPT}               &      \multicolumn{1}{c}{TDA}      \\ \midrule
        \textcolor{red}{\xmark}                 &         10.07             &   509.87              &       511.33              &     10.84     \\ 
    \textcolor{ForestGreen}{\omark}         &         10.16     &   510.22     &       511.55     &     11.27      \\ \cmidrule(lr){1-5}
    Rel.    &      +0.9\%      &      +0.07\%    &     +0.04\%    &    +4.0\% \\
    
    \bottomrule
    \end{tabular}
    \captionof{table}{\textbf{Per-sample cost analysis.} \alg introduces less than 0.5 ms of additional computational overhead across the evaluated TTA methods.}
    \label{tab:time_cost}
    \vspace{-0.4cm}
    
\end{minipage}
\end{table*}

\subsection{Further Analysis Results}

\begin{figure}[t]
    \begin{minipage}{0.53\textwidth}
        \includegraphics[width=\linewidth]{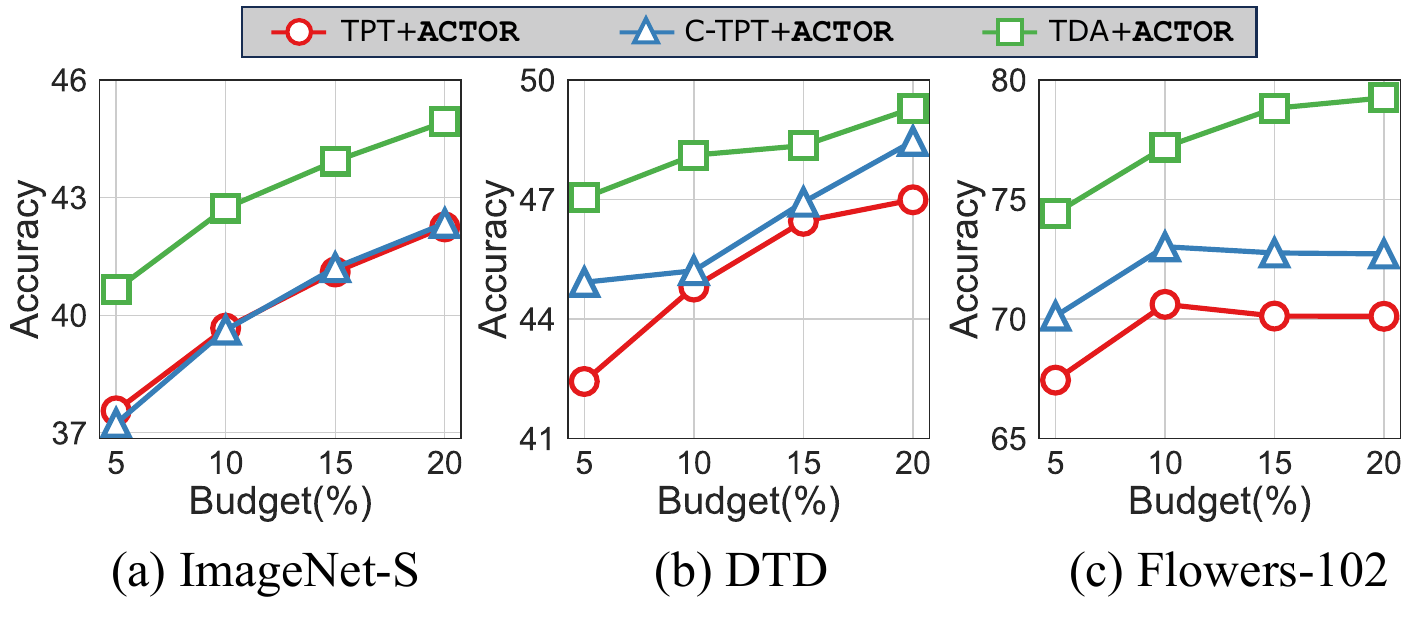}
        \vspace{-0.3cm}
        \caption{\textbf{Performance variation with budget across three datasets.}
        TDA+\alg achieves the best performance under varying budget constraints.
        }
        \vspace{-0.3cm}
        \label{fig:budget}
    \end{minipage}
    \hfill
    \begin{minipage}{0.45\textwidth}
        \centering
        \vspace{-0.3cm}
        \includegraphics[width=1.0\linewidth]{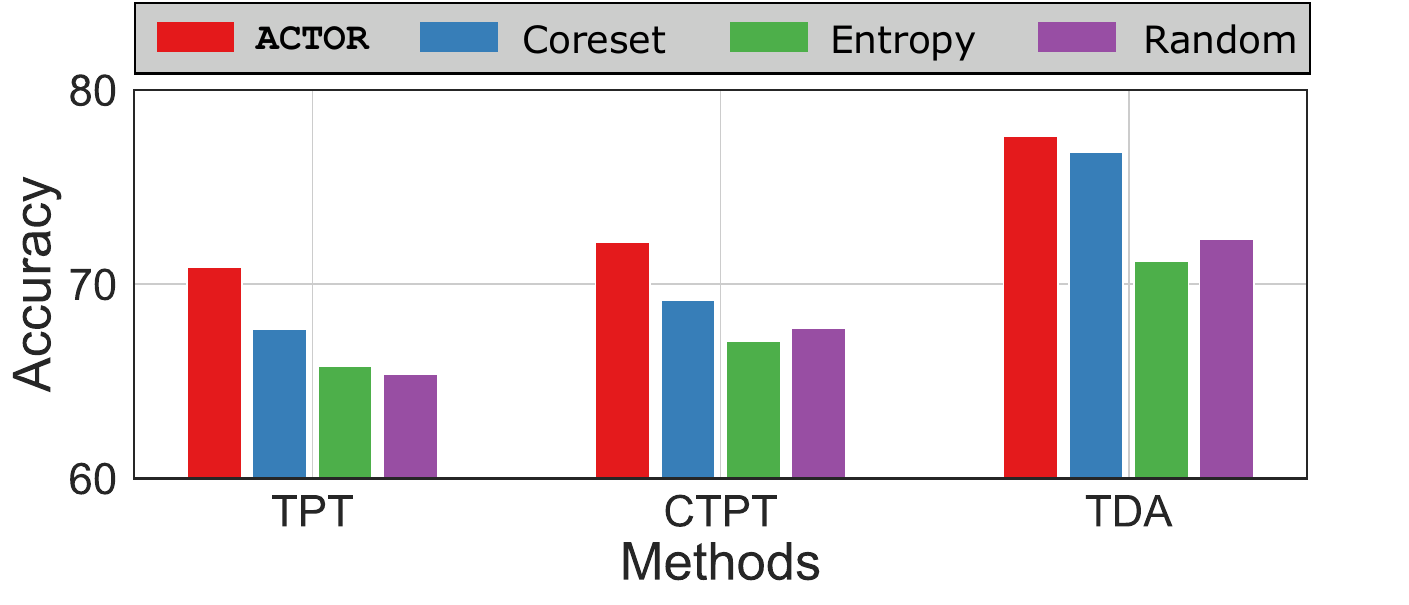}
        \vspace{-0.7cm}
        \caption{\textbf{Comparison of conventional sample selection methods on Flowers-102.} All the active learning methods are adapted to the stream setup (see Appendix~\ref{app:baselines}).}
        \vspace{-0.3cm}
        \label{fig:al}
        \vspace{-0.7cm}
    \end{minipage}
\end{figure}

\myparagraph{Performance across label budgets.}
\alg's retrieval relies on the labeled memory $\mathfrak{D}_l(t)$, so a larger budget enlarges this memory and enables more reliable retrieval. Figure~\ref{fig:budget} shows that the gains are not confined to the 10\% endpoint used in our main tables: across all three datasets, \alg{} improves throughout the budget sweep, indicating that queried supervision is useful before the final budget is exhausted and that additional labels continue to sharpen retrieval.

\setlength{\tabcolsep}{15pt}

\begin{figure}[t] 
    \centering
    \begin{minipage}{0.48\linewidth}
        \centering
        \begin{subfigure}[b]{0.48\linewidth}
            \centering
            \includegraphics[width=\textwidth]{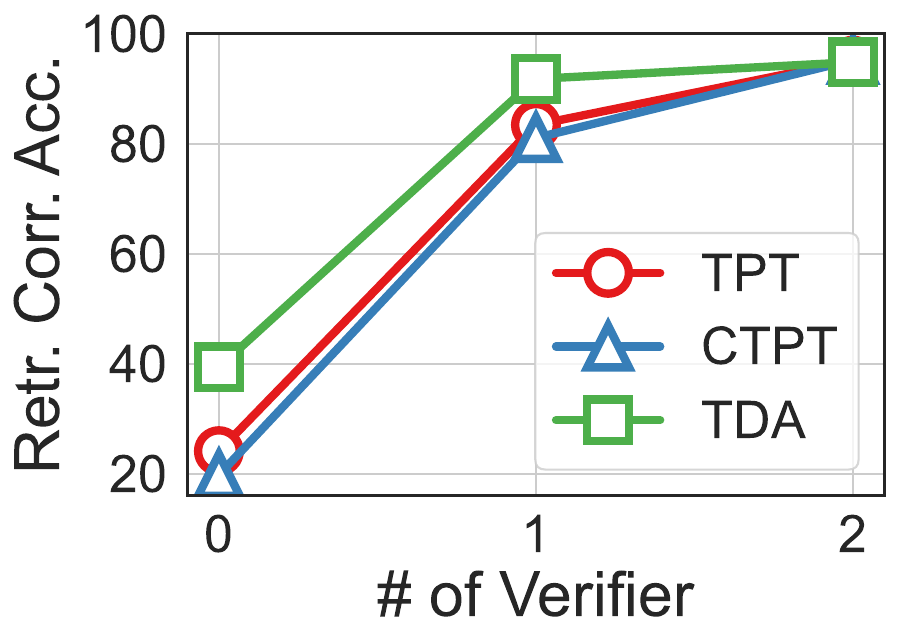}
            \vspace{-0.5cm}
            \caption{}
            \label{fig:k_reuse_acc}
        \end{subfigure}
        \hfill
        \begin{subfigure}[b]{0.46\linewidth}
            \centering
            \includegraphics[width=\textwidth]{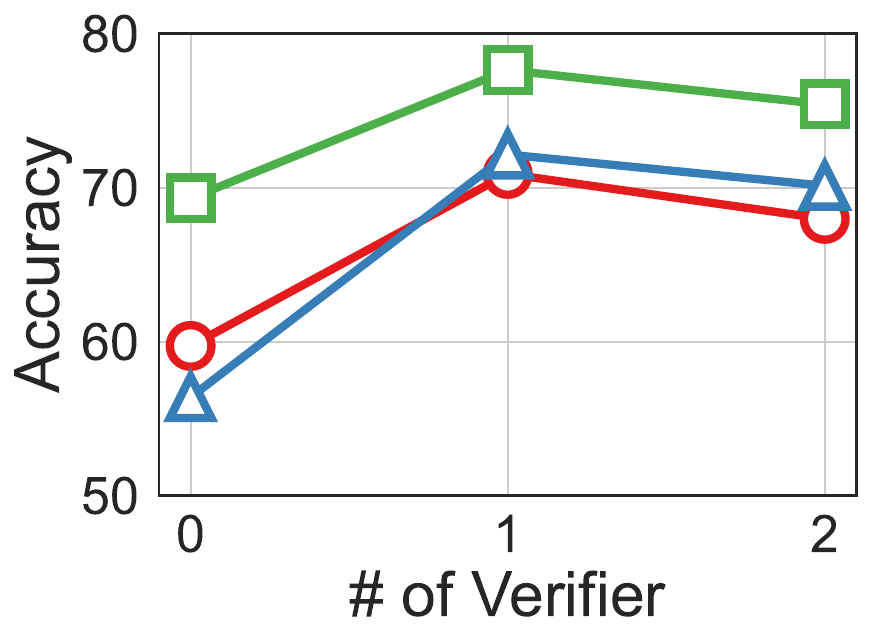}
            \vspace{-0.5cm}
            \caption{}
            \label{fig:k_acc}
        \end{subfigure}
        \vspace{-0.25cm}   
        \caption{\textbf{Performance changes as the number of Verifiers increases.} (a) indicates the accuracy specifically for retrieval-augmented correction, whereas (b) refers to the accuracy calculated over all samples.}
        \label{fig:K}
    \end{minipage}
    \hfill
    \begin{minipage}{0.48\linewidth}
        \centering
        \includegraphics[width=1.0\linewidth]{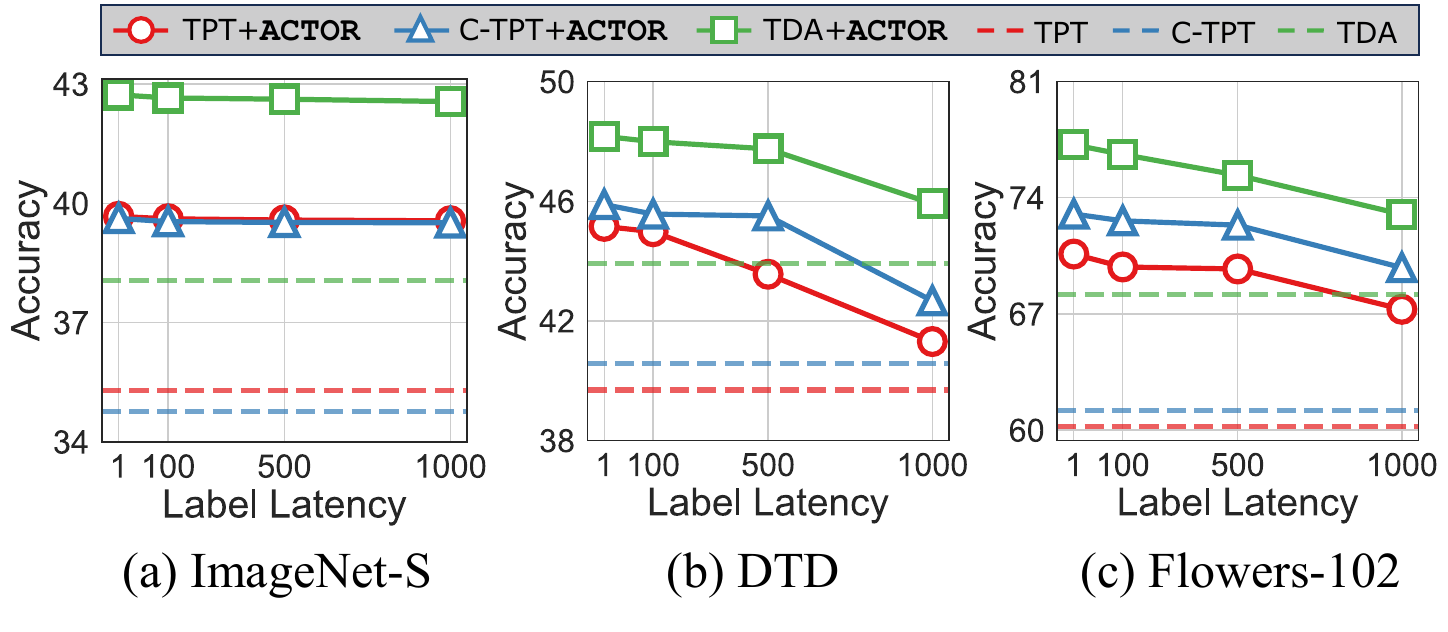}
        \vspace{-0.7cm}
        \caption{\textbf{Robustness to label latency in \alg.} \emph{Labeling latency} is defined as the number of test samples that elapse between the moment \alg{} decides to query $x_t$ and the moment the oracle returns $y_t$. 
        }
        \vspace{-0.3cm}
        \label{fig:label_latency}
    \end{minipage}
\end{figure}

\myparagraph{Comparison with diverse sample selection methods.} 
We compare \alg with standard streaming query-selection strategies---Random, Coreset~\citep{sener2018active}, and Entropy~\citep{holub2008entropy}---under the same labeling budget (See Figure~\ref{fig:al}). To isolate the effect of which samples are queried, all baselines use the same delayed-label protocol and the same retrieval memory once queried labels become available; they differ only in the criterion used to select samples for annotation. Random selects queried samples uniformly, Coreset prioritizes samples that are far from the current labeled memory in the embedding space, and Entropy prioritizes samples with high predictive entropy (See Appendix~\ref{app:baselines}).

\myparagraph{Impact of a Verifier $(x_t^V, y_t^V)$.} Verifiers minimize retrieval-augmented correction errors (Section~\ref{subsec:retr_pred}). Varying the number of Verifiers in Figure~\ref{fig:K}, we observe that introducing a Verifier ($0\!\to\!1$) significantly boosts correction accuracy (Figure~\ref{fig:K}a), with further gains as Verifiers increase. However, Figure~\ref{fig:K}b shows overall accuracy dips when $1\!\to\!2$ because Equation~\ref{eq:criteria} becomes overly strict, reducing the number of corrected samples.



\myparagraph{Sensitivity of labeling latency of \alg.} In this analysis, we explicitly vary $\Delta_t$ to test delayed annotation.
Some delay in human labeling is unavoidable. Figure~\ref{fig:label_latency} shows that longer delays gradually reduce the performance gap of \alg over TTA, yet \alg\ still matches TTA's performance at worst. We test delays of $\{1, 100, 500, 1000\}$ samples on ImageNet-S (50k), DTD ($\sim$2k), and Flowers-102 ($\sim$2k). On the large ImageNet-S, performance is stable (around 43\% for TDA+\alg and 40\% for C-TPT+\alg). 
On the smaller DTD and Flowers-102, the same delays produce larger drops (\eg DTD: $46\%\!\rightarrow\!42\%$)
, confirming that degradation occurs when latency constitutes a significant fraction of the stream; larger streams mitigate this effect.

\begin{figure}[t]
    \centering
    \includegraphics[width=0.85\linewidth]{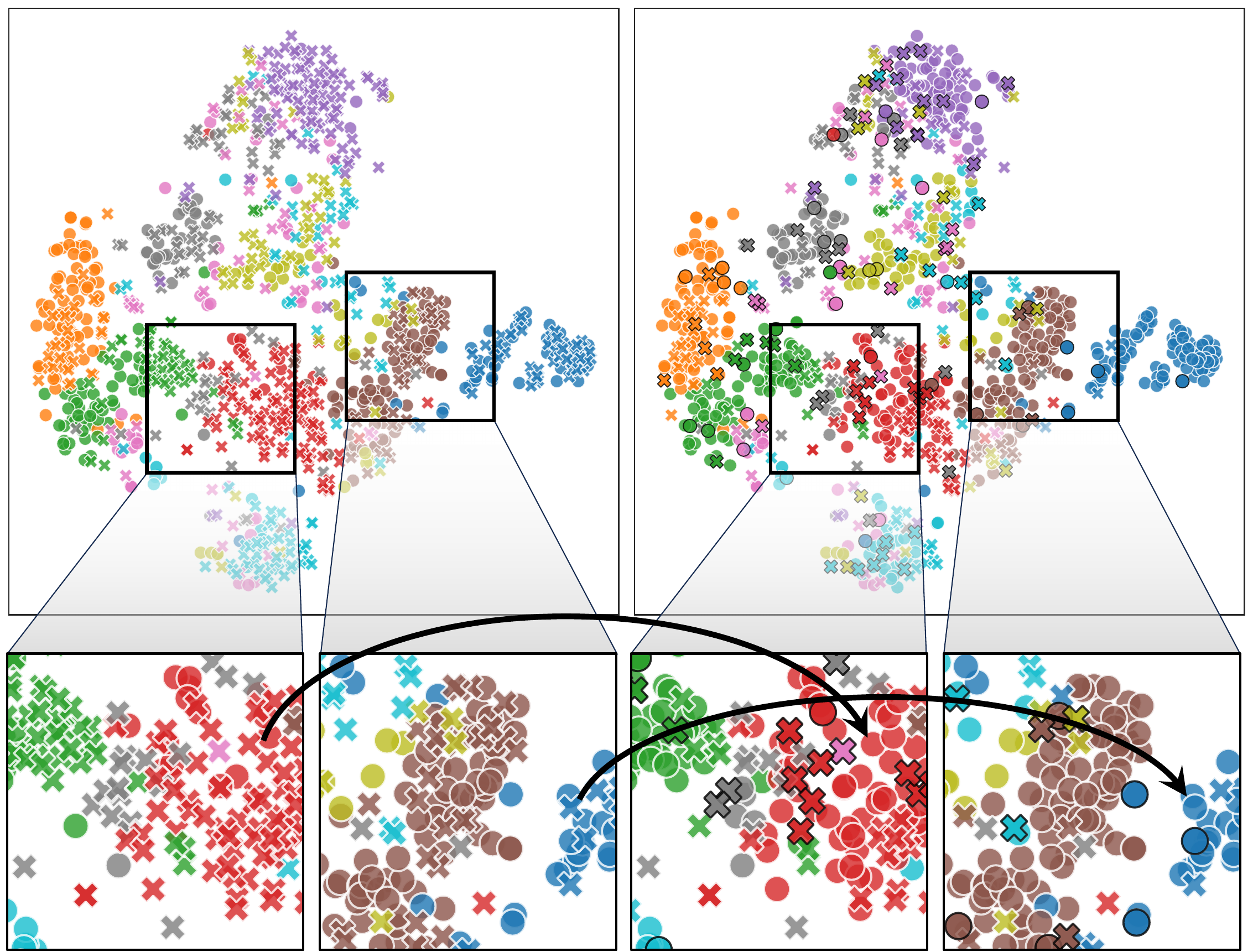}
    \caption{
    \textbf{Visualization of retrieval-augmented prediction correction on EuroSAT.} Circles ($\circ$) indicate correctly classified samples, while crosses ($\times$) indicate misclassified samples. Black outlines denote samples queried and labeled by \alg.
    }
    \label{fig:tsne_comp}
    \vspace{-0.3cm}
\end{figure}

\myparagraph{\alg{} corrects predictions in local embedding neighborhoods.} Figure~\ref{fig:tsne_comp} illustrates how \alg corrects misclassified samples using labeled anchors in local neighborhoods of the visual embedding space: the examples with black outlines (selected Referrers/Verifiers) exert noticeable influence in their neighborhoods, where many initially misclassified samples (crosses in Figure~\ref{fig:tsne_comp}(left)) are corrected to align with the ground truth (circles in Figure~\ref{fig:tsne_comp}(right)).

\myparagraph{Conservative reuse reduces harmful corrections.}
Table~\ref{tab:harm_main} isolates the effect of different label reuse rules by fixing the test stream, queried samples, returned labels, labeling budget, and annotation latency across all methods, while varying only the retrieval-based correction rule.
Using TDA with CLIP-ViT-B/16, we evaluate average accuracy across Flowers102, DTD, Caltech101, IN-S, and IN-V2. The full consistency-and-reliability rule achieves 69.50\% accuracy, improving upon base TDA (65.31\%) by 4.19 percentage points (pp) and the strongest plain kNN baseline, 5-NN majority voting (67.22\%), by 2.28 pp. It also reduces the harmful-correction rate, defined as the fraction of applied corrections that turn a correct base prediction into an incorrect one (13.68\% $\rightarrow$ 5.28\%), an 8.40 pp reduction relative to 5-NN majority voting. Consistency-only reuse improves accuracy to 68.78\% but retains a relatively high harm rate of 11.99\%, while reliability-only reuse yields limited improvement (67.30\% accuracy and 13.56\% harm). 

\myparagraph{Hyperparameter sensitivity.}
\alg{} introduces two hyperparameters, $\tau_u$ and $\tau_r$, that decide whether a sample should undergo retrieval-augmented correction. First, $\tau_u$ dictates how many samples rely on TTA predictions. 
Specifically, lowering $\tau_u$ makes more samples appear \emph{uncertain}, diverting them away from direct TTA-based decisions, whereas raising $\tau_u$ increases the likelihood of accepting TTA outputs. 
Second, $\tau_r$ governs which of these uncertain samples undergo retrieval-augmented correction. A lower $\tau_r$ includes more samples in retrieval-augmented correction, while a higher $\tau_r$ imposes stricter conditions, causing more samples to label.

To assess potential sensitivity of \alg{}, we tested different values of $\tau_u$ and $\tau_r$, as reported in Figure~\ref{fig:hyperparams}. Results across three datasets show that each one behaves differently due to varying levels of difficulty in each dataset. Nonetheless, on every dataset and for every TTA method, incorporating \alg{} consistently surpasses the baseline (\ie \emph{without} \alg{}), regardless of the specific hyperparameter settings. This observation confirms the robustness for a reasonable range of the thresholds.

%% file: revised_sections/05_related.tex
\section{Related Work}
\label{sec:related}

\myparagraph{Active learning.} 
Active learning aims to minimize labeling cost by selecting informative samples~\citep{settles2009active,ren2021survey}, using uncertainty-based criteria~\citep{lewis1994heterogeneous,roth2006margin,holub2008entropy}, diversity-based criteria~\citep{sener2018active,parvaneh2022active}, or hybrids such as BADGE~\citep{ash2019deep}. PCB~\citep{bang2024active} specializes active learning to VLMs, while online active learning~\citep{qin2021active,desalvo2021online} often still assumes continual retraining; in both cases, queried labels mainly serve to update a model or improve a final pool-trained learner. Active test-time adaptation is closer in spirit because it allows limited human feedback during deployment, but existing methods remain update-centric: for example, SimATTA~\citep{gui2024active} uses queried labels for batch-level online parameter updates, so labels affect future predictions only after annotation and a subsequent update step. In contrast, \alg{} keeps the VLM frozen and stores each returned label in memory as soon as it arrives, enabling retrieval-augmented correction for subsequent samples while explicitly modeling human label delay.

\myparagraph{Vision-language models.} 
Large-scale image--text pretraining has produced a family of VLMs including CLIP~\citep{radford2021learning}, ALIGN~\citep{jia2021scaling}, BLIP/BLIP-2~\citep{li2022blip,li2023blip}, LiT~\citep{zhai2022lit}, SigLIP~\citep{zhai2023sigmoid}, and FILIP~\citep{yao2021filip}, which align images and text in a shared embedding space and support zero-shot classification. Beyond zero-shot use, prior work has adapted VLMs with offline labeled support through prompt learning methods such as CoOp~\citep{zhou2022learning}, CoCoOp~\citep{zhou2022conditional}, and MaPLe~\citep{khattak2023maple}, as well as feature/cache-based adapters such as CLIP-Adapter~\citep{gao2023clip} and Tip-Adapter~\citep{zhang2022tip}. These methods assume a labeled support set is already available before deployment and then optimize how to fit or reuse it. \alg\ exploits the same shared-space property for retrieval, but differs in that its labeled memory is not given in advance and cannot be used to revise the queried sample itself: it is built on the fly from delayed deployment-time queries under a streaming budget.

\myparagraph{Test-time adaptation for VLMs.} 
To bridge the pretraining--deployment gap, TPT~\citep{shu2022tpt} tunes prompts per test sample via augmentation-based entropy minimization, DiffTPT~\citep{feng2023difftpt} augments the input with diffusion-generated views, and C-TPT~\citep{yoon2024ctpt} calibrates the prediction uncertainty of TPT. RLCF~\citep{zhao2023rlcf} observes that entropy-based losses can get stuck on wrong predictions and proposes a CLIP-score-based objective. Moving away from per-sample optimization, TDA~\citep{karmanov2024efficient} introduces training-free TTA via Tip-Adapter~\citep{zhang2022tip}-style caching, reducing inference cost while still adapting online. These methods are label-free and are therefore complementary to our setting. More broadly, retrieval-based adaptation has also been studied outside our setting~\citep{zancato2023train, lee2025ra}, but these methods do not model label acquisition from the deployed stream under an explicit query budget. 

\begin{figure}[t]
    \centering
    \scriptsize
    \begin{minipage}{0.48\textwidth}
        \centering
        \setlength{\tabcolsep}{10pt}
        \begin{tabular}{lcc}
        \toprule
        \scriptsize
        \textbf{Reuse rule} & \textbf{Avg.} ($\uparrow$) & \textbf{Harm.} ($\downarrow$) \\
        \midrule
        Base TDA & 65.31{\tiny $\pm$0.26} & -- \\ \midrule
        Hard 1-NN & 67.00{\tiny $\pm$0.38} & 14.98{\tiny $\pm$0.96} \\
        5-NN majority & 67.22{\tiny $\pm$0.28} & 13.68{\tiny $\pm$0.04} \\
        Weighted 5-NN & 67.21{\tiny $\pm$0.28} & 13.69{\tiny $\pm$0.06} \\ \midrule
        Consistency only & 68.78{\tiny $\pm$0.23} & 11.99{\tiny $\pm$0.98} \\
        Reliability only & 67.30{\tiny $\pm$0.34} & 13.56{\tiny $\pm$0.67} \\
        \rowcolor[gray]{.9} \alg{} (both) & \textbf{69.50}{\tiny $\pm$0.32} & \textbf{5.28}{\tiny $\pm$0.26} \\
        \bottomrule
        \end{tabular}%
        \captionof{table}{\textbf{Effect of correction reuse rules under a shared memory.} Average accuracy (Avg.) and harmful-correction rate (Harm.) across five datasets using TDA with CLIP-ViT-B/16. All methods share the same queried samples and returned labels under an identical labeling budget and annotation latency, differing only in their label reuse rules. Harm. measures the fraction of applied corrections that change a correct base prediction into an incorrect one. }
        \label{tab:harm_main}
    \end{minipage}
    \hfill 
    \begin{minipage}{0.48\textwidth}
        \vspace{-0.5cm}
        \centering
        \includegraphics[width=\linewidth]{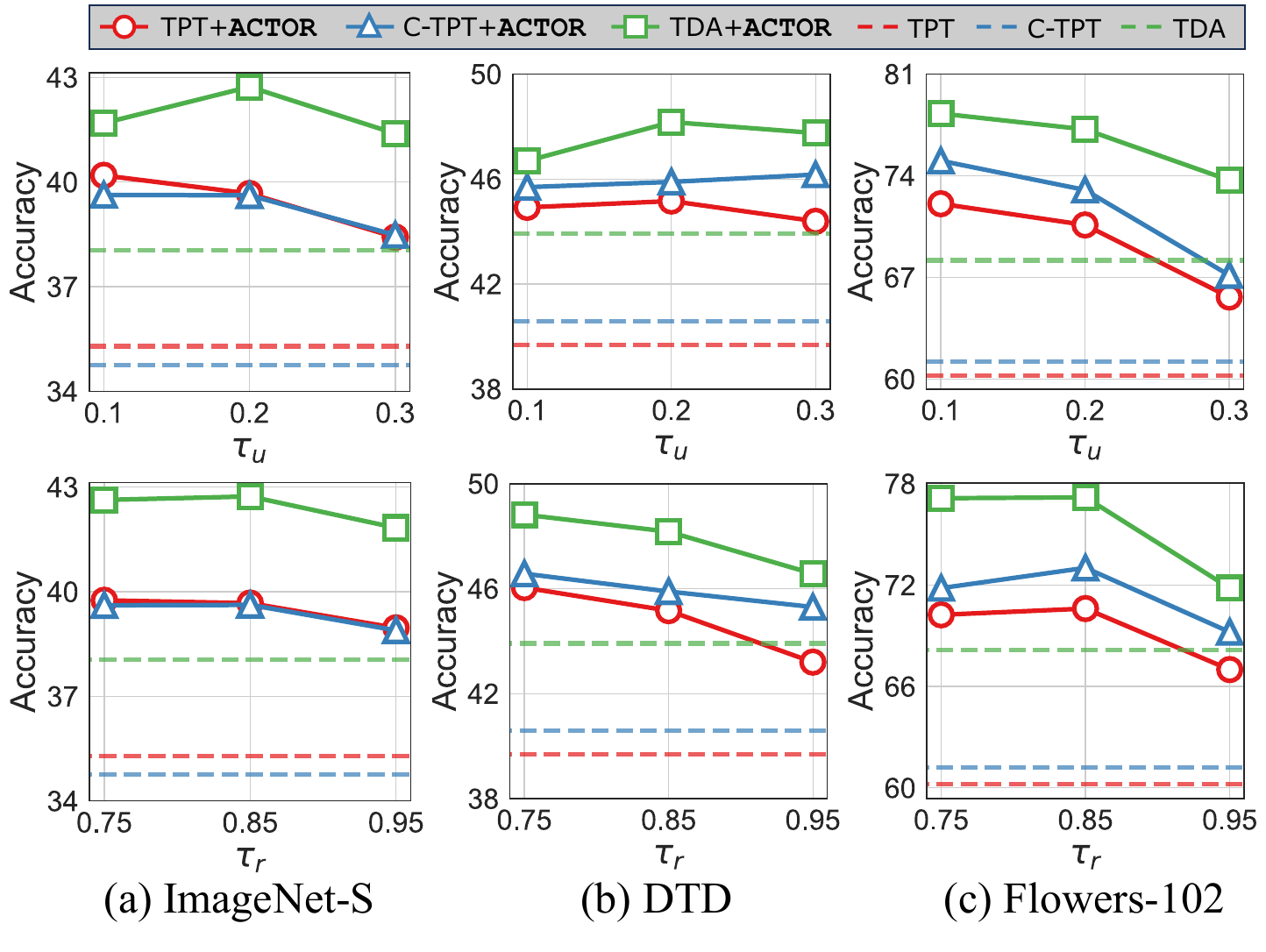}
        \vspace{-0.3cm}
        \captionof{figure}{\textbf{Hyperparameter sensitivity.} We vary the uncertainty threshold $\tau_u$ with $\tau_r=0.85$ (upper) and the retrieval-augmented correction threshold $\tau_r$ with $\tau_u=0.2$ (lower).}
        \label{fig:hyperparams}
    \vspace{-0.5cm}            
    \end{minipage}
    \vspace{-0.3cm}    
\end{figure}

%% file: revised_sections/06_Conclusion.tex
\section{Conclusion}
\label{sec:conclusion}

We introduced \textbf{Active VLM-TTA}, a streaming protocol between fully unsupervised TTA and supervised online learning. In this setting, a frozen VLM may spend an explicit label budget, but queried labels are delayed and can only improve future predictions. We instantiated this protocol with \alg, a training-free plug-in framework that reuses sparse queried labels through retrieval-augmented correction. Across OOD and cross-domain benchmarks, \alg\ consistently improves strong VLM-TTA baselines on both ResNet and ViT backbones with minimal computational overhead, and a label-matched CoOp comparison shows that the gains are not merely due to label access.

\section*{Limitations}
\label{sec:limitation}
Active VLM-TTA is intended to complement, rather than replace, label-free TTA, because it explicitly models deployment scenarios where limited delayed human feedback is available. \alg is designed for settings in which queried annotations are reliable and the labeled memory can provide local evidence for future samples. Although our latency and budget analyses show that the method remains robust across the tested ranges, its effectiveness may decrease when the returned labels are noisy, the stream provides highly uneven class coverage, or the annotation budget is exhausted and correction must rely more heavily on Top-$K$ plausibility. Extending \alg with noise-aware memory filtering, class-balanced querying, distance-aware fallback, and stream-order-aware memory management is an important direction for future work.

%% file: revised_sections/07_ACK.tex
\section*{Acknowledgment}
This work was supported by Institute of Information \& Communications Technology Planning \& Evaluation\,(IITP) grant funded by the Korea government\,(MSIT) (No.\ RS-2020-II200862, DB4DL: High-Usability and Performance In-Memory Distributed DBMS for Deep Learning, 40\%, No.\ RS-2022-II220157, Robust, Fair, Extensible Data-Centric Continual Learning, 40\%, No.\ RS-2026-25614738, AI Star Fellowship Support Program, 5\%, and No.\ RS-2025-25464461, AI's Vision of Harmony: A Fair and Transparent Multimodal Agentic Platform for Conflict Mediation, 5\%) and the InnoCORE program of the Ministry of Science and ICT (AI Meta-Scientist, N10260110, 10\%).

%% file: revised_sections/10_Appendix.tex
\section{Conventional TTA}
\label{app:tta}
Test-time adaptation (TTA) using vision-language models (VLMs) can be classified into two types: prompt tuning-based approaches (\eg TPT~\citep{shu2022tpt} and C-TPT~\citep{yoon2024ctpt}) and training-free approaches (\eg TDA~\citep{karmanov2024efficient}). Depending on the approach, we can formulate the TTA operator $\mathcal{T}_\xi(\cdot)$ as follows. Note that C-TPT is omitted here due to its conceptual similarity to TPT.

\subsection{Test-time Prompt Tuning (TPT)}
TPT is achieved by fine-tuning a learnable prompt $p$ using a single test sample $x_t$. Given multiple augmented views of $x_t$, denoted by $A_i(x_t)$ for $i=\{1,\dots,M\}$, the goal is to minimize the entropy (as defined in Equation~\ref{eq:entropy}) of the model's predictions across these augmentations.

\myparagraph{Objective Function.}
The probability distribution over classes generated by the CLIP model with prompt \(p\) on the \(i\)-th augmented view can be formulated as
\begin{equation*}
    p^{(i)}(x_t) = p(A_i(x_t)) = \frac{\exp(\mathbf{s}^{(i)} / \tau_{\rm temp})}{\sum_{k=1}^{\mathcal{C}}{\exp(s_k^{(i)} / \tau_{\rm temp})}},
\end{equation*}
where $\mathbf{s}^{(i)} = \{s_c^{(i)} | c \in \{1, \ldots, \mathcal{C}\}\}$, $\tau_{\rm temp}$ denotes the temperature, and the similarity $ s_c^{(i)}$ is defined as
\begin{equation*}
    s_c^{(i)} = \cos\big(f_{\text{img}}(A_i(x_t)), f_{\text{txt}}([p;t_c])\big),
\end{equation*}
where $t_c$ is the text prompt of class $c$. Note that it is similarly calculated by Equation~\ref{eq:cos_sim}, but the difference is that the text prompt concatenates learnable prompt $p$ and text prompt $t_c$ for class $c$, and the input image is augmented when calculating the similarity score. Then, the averaged prediction $\tilde{p}_p(x_t)$ (with confidence selection) is defined as

\[
\tilde{p}_p(x_t)=\frac{1}{\rho M}\sum_{i=1}^{M}
\mathbf{1}\left[H(p^{(i)}(x_t))\le \tau_{\rm ent}\right]p^{(i)}(x_t),
\]%
where $\mathcal{H}(\cdot)$ denotes the entropy function, $\tau_{\rm ent}$ is the entropy threshold, and $\rho$ is the proportion of augmented views selected.

The optimization problem is then given by
\begin{equation*}
p^* = \arg\min_{p} \mathcal{L}_{\text{TPT}}(p;x_t),
\end{equation*}
with the loss defined as the marginal entropy,
\begin{equation*}
\mathcal{L}_{\text{TPT}}(p;x_t) = -\sum_{y\in\mathcal{Y}} \tilde{p}_{p,y}(x_t)\log \tilde{p}_{p,y}(x_t).
\end{equation*}

After optimization, the test-time adapted prediction function is defined as
\begin{equation*}
    \mathcal{T}_\xi(x_t) = \mathcal{T}_{p^*}(x_t) \triangleq p_{p^*}(x_t).
\end{equation*}

\subsection{Training-free Dynamic Adapter (TDA)}
In contrast to prompt tuning-based approaches, TDA leverages a non-parametric dynamic cache to adapt predictions without backpropagation. Here, $\xi$ comprises the components of the dynamic cache,
\begin{equation*}
\xi = \{Q_p,\, \hat{L}_p,\, Q_n,\, \hat{L}_n\},
\end{equation*}
where $(Q_p, \hat{L}_p)$ denotes the keys and corresponding pseudo labels in the \textit{positive cache} (collected from high-confidence predictions), and $(Q_n, \hat{L}_n)$ denotes the keys and corresponding negative pseudo labels in the \textit{negative cache} (collected from low-confidence predictions).

Then, the final prediction is computed as
\begin{equation*}
\begin{split}
\mathcal{T}_\xi(x_t) = &f_{\mathrm{img}}(x_t) W_T^c + \mathcal{A}\big(f_{\mathrm{img}}(x_t) Q_p^T\big)\,\hat{L}_p - \mathcal{A}\big(f_{\mathrm{img}}(x_t) Q_n^T\big)\,\hat{L}_n,
\end{split}
\end{equation*}
where $W_T^c = \big[f_{\mathrm{txt}}(t_c) \mid c\in\mathcal{C}\big]$ represents the text embedding matrix computed from the class names, and $\mathcal{A}(\cdot)$ is an adapter function (\eg an exponential weighting function as used in Tip-Adapter~\citep{zhang2022tip}).

\section{Algorithm Details}
\label{app:alg}

Algorithm~\ref{alg:atta} outlines our proposed method, \alg. For each test sample $x_t$, our method first assesses its prediction uncertainty. If the sample is certain, \alg directly returns the base TTA prediction. For uncertain samples, it employs a hierarchical strategy to manage its explicit labeling budget. The process begins with a warm-up phase, during which uncertain samples are queried until at least $N$ returned labels are available in the labeled memory $\mathfrak{D}_l$, subject to the labeling budget $\mathfrak{B}$. Pending annotations are counted toward the budget but are incorporated into memory only after the specified latency. Importantly, each prediction is emitted before its annotation is requested, ensuring that queried labels cannot revise the predictions of their corresponding samples.

Once sufficient labeled samples are available, \alg's core \textbf{retrieval-augmented correction} mechanism is activated. It first attempts a budget-free correction by retrieving labeled anchors from $\mathfrak{D}_l$ and checking whether their labels are consistent and their features are sufficiently close to the current sample. If the correction criterion is not satisfied and the remaining labeling budget permits, the algorithm queries the oracle for future use. After budget exhaustion, it instead applies the Top-$K$ plausibility check as a fallback when the original correction criterion is not satisfied. This strategy enables conservative label reuse while preserving the future-only constraint of Active VLM-TTA.

\begin{algorithm}[h]
    \DontPrintSemicolon
    \SetAlgoLined
    \SetNoFillComment
    \LinesNotNumbered
    \caption{\alg}
    \label{alg:atta}
    \KwIn{
        Labeled memory $\mathfrak{D}_l(t)$,
        Pending annotations $\mathcal{P}(t)$,
        Number of issued queries $Q(t)$,
        Uncertainty threshold $\tau_u$,
        Latency $\Delta$,
        Correction threshold $\tau_r$,
        Labeling budget $\mathfrak{B}$,
        Minimum labeled samples $N$,
        TTA method $\mathcal{T}_{\xi}(\cdot)$,
        Human annotator $\texttt{Human}(\cdot)$,
        Test sample $x_t$
    }

    Incorporate returned annotations from $\mathcal{P}(t)$
    into $\mathfrak{D}_l(t)$\;

    $q_t \gets
    \mathcal{T}_{\xi}(x_t; f_{\mathrm{img}}, f_{\mathrm{txt}})$\;
    Normalize $q_t$ with softmax if needed\;
    $u \gets \mathcal{H}(q_t)$ from Equation~\ref{eq:entropy}

    $\hat{p}_t \gets q_t$\;
    $\mathrm{query} \gets \mathrm{False}$\;

    {\small{\textcolor{Skyblue}{
    \# \textbf{Uncertain sample}: Apply \alg}}}\\

    \If{$u \ge \tau_u$}{

        {\small{\textcolor{Skyblue}{
        \# \textbf{Warm-up}: Construct labeled memory}}}\\

        \uIf{$|\mathfrak{D}_l(t)| < N$}{
            \If{$Q(t) < \mathfrak{B}$}{
                $\mathrm{query} \gets \mathrm{True}$\;
            }
        }
        \Else{
            Retrieve $\mathrm{R},\mathrm{V}
            \in \mathfrak{D}_l(t)$\;

            Compute $\mathcal{C}(x_t|\mathrm{R},\mathrm{V})$ from Equation~\ref{eq:criteria}

            {\small{\textcolor{Skyblue}{
            \# Retrieval-augmented correction
            $\triangleright$~Section~\ref{subsec:retr_pred}}}}\\

            \uIf{$\mathcal{C}(x_t|\mathrm{R},\mathrm{V})
            > \tau_r$}{
                $\hat{p}_t \gets
                \mathtt{one\_hot}(y_t^R)$\;
            }
            \uElseIf{$Q(t) < \mathfrak{B}$}{
                $\mathrm{query} \gets \mathrm{True}$\;
            }
            \Else{
                {\small{\textcolor{Skyblue}{
                \# Correction after budget exhaustion
                $\triangleright$~Section~\ref{subsec:saturated}}}}\\

                Compute $\hat{p}_t$ using
                Equation~\ref{eq:budget}\;
            }
        }
    }

    {\small{\textcolor{Skyblue}{
    \# Emit prediction before requesting annotation}}}\\
    $\texttt{Emit}(\hat{p}_t)$\;

    \If{$\mathrm{query}$}{
        Request annotation $\texttt{Human}(x_t)$\;
        Add the request to $\mathfrak{P}(t)$
        with return time $t+\Delta$\;
        $Q(t+1) \gets Q(t)+1$\;
    }
\end{algorithm}

\section{Experimental Setup}
\label{app:exp}
\subsection{Datasets}
Our experiments were conducted on two benchmarks: the out-of-distribution (OOD) benchmark and the cross-domain benchmark, both used in previous research on adapting VLMs during test time. The \underline{OOD benchmark} evaluates the robustness of our approach on ImageNet~\citep{deng2009imagenet} and four ImageNet-derived distribution-shift datasets: ImageNet-A~\citep{hendrycks2021nae}, ImageNet-V2~\citep{recht2019imagenet}, ImageNet-R~\citep{hendrycks2021many}, and ImageNet-S~\citep{wang2019learning}. In contrast, the \underline{cross-domain benchmark} examines the model's adaptability across different domains by evaluating it on ten diverse image classification datasets, each representing a distinct class space. These datasets include Aircraft~\citep{maji2013fine}, Caltech101~\citep{fei2004learning}, Stanford Cars~\citep{krause20133d}, DTD~\citep{cimpoi2014describing}, EuroSAT~\citep{helber2019eurosat}, Flowers102~\citep{nilsback2008automated}, Food101~\citep{bossard2014food}, Oxford Pets~\citep{parkhi2012cats}, SUN397~\citep{xiao2010sun}, and UCF101~\citep{soomro2012ucf101}. Using these two benchmarks, we provide a comprehensive evaluation of the model's generalization capability and adaptability during test time. We summarize the statistics of datasets in Table~\ref{tab:dataset_statistic}, and the details are provided below.

\subsubsection{OOD Benchmarks}
\noindent \textbf{ImageNet}~\citep{deng2009imagenet} contains 14,197,122 annotated images according to the WordNet hierarchy. This dataset has been used in the ImageNet large scale visual recognition challenge (ILSVRC), a benchmark in image classification and object detection, since 2010.

\noindent \textbf{ImageNet-A}~\citep{hendrycks2021nae} is a subset of 7,500 visually similar but naturally perturbed ImageNet images of 200 classes.

\noindent \textbf{ImageNet-V2}~\citep{recht2019imagenet} consists of 10,000 images and 1,000 ImageNet classes, and was collected by applying an updated natural data collection pipeline to the original ImageNet dataset.

\noindent \textbf{ImageNet-R}~\citep{hendrycks2021many} includes 30,000 images belonging to 200 categories of the ImageNet dataset, but with diverse artistic styles.

\noindent \textbf{ImageNet-S}~\citep{wang2019learning} consists of 50,889 sketches of 1,000 class objects from the ImageNet dataset, and represents a domain shift from natural images to sketches.

\subsubsection{Cross Domain Benchmarks}
\noindent \textbf{FGVC-Aircraft}~\citep{maji2013fine} comprises a total of 10,000 images depicting various aircraft, organized into 100 distinct classes. Each class corresponds to a specific aircraft model variant and contains 100 images. The dataset is divided into training, validation, and test sets, with the test set comprising 3,333 images across all 100 classes. The class labels specify the aircraft manufacturer, model, and variant, such as Boeing 737-76J.

\noindent \textbf{Caltech101}~\citep{fei2004learning} originally comprises 9,144 images spanning 101 object categories and an additional background category. These categories encompass a wide range of objects, including animals, vehicles, and household items. Following previous evaluation protocols, we exclude the \texttt{BACKGROUND} and \texttt{Faces easy} categories, resulting in 100 classes. The test set consists of 2,465 images distributed unevenly across these classes. Notably, Caltech101 exhibits substantial class imbalance due to the varying numbers of images across its categories.

\noindent \textbf{Stanford Cars}~\citep{krause20133d} consists of a collection of 16,185 images categorized into 196 different classes, with each class typically representing specific car make, model, and year, \eg 2012 Tesla Model S.

\setlength{\tabcolsep}{20pt}
\begin{table}[t]
  \centering
  \scriptsize
  \begin{tabular}{@{}lrr@{}}
    \toprule
    Dataset         & \# of Classes   & \# of Test Instances \\ \cmidrule(lr){1-3}
    ImageNet        &       1000           &   50,000 \\
    ImageNet-A      &        200           &    7,500 \\
    ImageNetV2      &       1000           &   10,000 \\
    ImageNet-R      &        200           &   30,000 \\
    ImageNet-S      &       1000           &   50,889 \\ \cmidrule(lr){1-3}
    
    FGVC Aircraft   &        100           &     3,333 \\ 
    Caltech101      &        100           &     2,465 \\
    Stanford Cars   &        196           &     8,041 \\
    DTD             &         47           &     1,692 \\
    EuroSAT         &         10           &     8,100 \\
    Flowers102      &        102           &     2,463 \\
    Food101         &        101           &    30,300 \\
    Oxford Pets     &         37           &     3,669 \\
    SUN397          &        397           &    19,850 \\
    UCF101          &        101           &     3,783 \\
    
    \bottomrule
    \end{tabular}
    \caption{\textbf{Profiles of the datasets used for the experiments.}
  } 
  \label{tab:dataset_statistic}
\end{table}

\noindent \textbf{DTD}~\citep{cimpoi2014describing}, abbreviated from Describable Texture Dataset, is designed for texture classification task. This dataset consists of 47 distinct classes, including categories like fabrics and natural materials. In total, DTD comprises 5,640 samples. Notably, when examining the performance reported in the CLIP~\citep{radford2021learning}, it becomes evident that DTD poses a challenging problem for pre-trained CLIP models, as textures are not typical, easily recognizable objects.

\noindent \textbf{EuroSAT}~\citep{helber2019eurosat} comprises 10 distinct classes that represent various land use and land cover categories. In total, this dataset includes 27,000 satellite images, with 2,700 images allocated to each of the 10 classes. Notably, each class contains an equal number of images, ensuring a balanced distribution within the dataset.

\noindent \textbf{Flowers102}~\citep{nilsback2008automated} consists of 102 different categories of flowers, each representing a distinct flower species such as roses, sunflowers, and daisies. There are 8,189 image-label pairs in total. Some categories have more images than the others, which means that it is imbalanced as typical real-world datasets; each category contains at least 40 and at most 258 samples.

\noindent \textbf{Food-101}~\citep{bossard2014food} consists of 101 food categories with 750 training and 250 test images per category, summing up to 101K images. The labels for the test images have been manually cleaned, while the training set contains some noise.

\noindent \textbf{Oxford Pets}~\citep{parkhi2012cats} consists of 37 different pet categories, including various dogs and cats. This dataset contains 7,400 samples. In particular, it has 4,978 dog images and 2,371 cat images. We use only class labels even though the dataset has segmentation, \ie both RoI and class.

\noindent \textbf{SUN397}~\citep{xiao2010sun} is the Scene UNderstanding (SUN) database that contains 899 categories and 130,519 images. There are 397 well-sampled categories to evaluate numerous state-of-the-art algorithms for scene recognition.

\noindent \textbf{UCF101}~\citep{soomro2012ucf101} is an extension of UCF50 and consists of 13,320 video clips, which are classified into 101 categories. These 101 categories can be classified into 5 types (body motion, human-human interactions, human-object interactions, playing musical instruments, and sports). The total length of these video clips is over 27 hours. All the videos are collected from YouTube and have a fixed frame rate of 25 FPS with the resolution of 320$\times$240. In this work, the middle frame of each video is fed to the image encoder.

\subsection{Baselines}
We compare \alg against representative strong TTA methods for VLMs, which can be categorized into prompt-tuning-based (\eg TPT, C-TPT) and training-free approaches (\eg TDA).

\myparagraph{Prompt tuning-based TTA.} We select TPT~\citep{shu2022tpt} and its concurrent extension C-TPT~\citep{yoon2024ctpt} as representatives of this category. These methods adapt a set of learnable prompt vectors for each individual test sample. The optimization is guided by a self-supervised objective that minimizes the entropy of the model's predictions across multiple augmented views of that sample, thereby fine-tuning the prompt without requiring true labels.

\myparagraph{Training-free TTA.} As a training-free counterpart, we use TDA~\citep{karmanov2024efficient}. In contrast to prompt tuning, TDA is a non-parametric method that requires no backpropagation at test time. It maintains a dynamic cache that stores features from past predictions, separating them into positive and negative caches using confidence scores. These caches are then used to adjust the prediction scores of the current test sample, effectively adapting the model on the fly.

\subsection{Implementation Details}
\label{app:implementation_detail}
 All models in our main experiments are built upon the pre-trained CLIP model, which consists of an image encoder and a text encoder. The image encoder can be either a ResNet or a Vision Transformer (ViT), while the text encoder is a Transformer. TTA is performed in a single-image setting with a batch size of 1. Specifically, the thresholds of uncertainty (See Equation~\ref{eq:entropy}) and retrieval-augmented correction (See Equation~\ref{eq:criteria}) are set to 0.2 and 0.85, respectively, the initial number of labeled examples $N$ is set to 5, and the explicit label budget $\mathfrak{B}$ is set to 10\% of the test stream. Queried labels are inserted into memory only for future samples. For evaluation, we use the \emph{top-1 accuracy} (\%) as the standard classification metric. 
 All experiments were repeated three times with different test sample orders generated using three fixed random seeds. We report the mean accuracy and standard deviation across the three runs. All experiments were conducted on a single NVIDIA RTX 4090.

\myparagraph{Hyperparameter protocol (validation-free).}
A common concern in the test-time adaptation literature is that strong results may arise from per-dataset hyperparameter tuning, which is not realistic under a streaming deployment assumption. We take the position that, by definition, Active VLM-TTA should not assume access to a per-task labeled validation split: the test stream itself is the only signal available at deployment, and the labeling budget is explicitly accounted for as a queried cost. We therefore fix a single set of hyperparameters $(\tau_u,\tau_r,N,\mathfrak{B}/|\mathfrak{D}_{\mathrm{test}}|)=(0.2, 0.85, 5, 10\%)$ \emph{before} seeing any benchmark-specific performance and use exactly this configuration on all 15 datasets (5 OOD + 10 cross-domain), both ResNet-50 and ViT-B/16 backbones, and all baselines (Zero-shot, TPT, C-TPT, TDA). The sensitivity study in Figure~\ref{fig:hyperparams} then shows that, for any choice within a reasonable range around these defaults, \alg{} still consistently outperforms the underlying TTA baseline, which suggests that the reported gains are not an artifact of a narrow, carefully tuned operating point.


\section{Further Analyses}
\label{app:further_analyses}
\subsection{Various VLMs}
\label{app:various_vlms}
While recent vision-language models (VLMs) such as BLIP-2~\citep{li2023blip} and SigLIP~\citep{zhai2023sigmoid} have emerged, the majority of prior TTA research has been conducted on CLIP. Therefore, for baseline consistency and fair comparison, we adopted CLIP in our main experiments (as presented in Table~\ref{tab:ood_main} and Table~\ref{tab:cd_main}). To further evaluate the generalizability of \alg, we extend our experiments to BLIP-2 and SigLIP, using visual features extracted from their respective pretrained image encoders for retrieval-based correction without additional training. As shown in Table~\ref{tab:various_vlms}, \alg consistently improves performance across these architectures, demonstrating its applicability to different VLM families.

\setlength{\tabcolsep}{10pt}
\begin{table}[t]
  \centering
  \scriptsize
  \begin{tabular}{lccccccc}
  \toprule
  \textbf{Model}  &  Aircraft    &    Caltech101    &    Cars    &    DTD    &    EuroSAT    &    IN      &    IN-A \\ \cmidrule(lr){1-8}
  CLIP            &   22.59      &     85.76        &    65.61   &   40.60   &    44.25      &    63.52   &    54.65 \\
  \rowcolor[gray]{.9} $~~$ + \alg &    \textbf{27.33}  &  \textbf{90.31}  &   \textbf{67.77}   & \textbf{47.94}   & \textbf{80.63}  & \textbf{65.90}  & \textbf{55.85} \\\cmidrule(lr){1-8}
  BLIP-2          &   12.75      &     92.58        &    78.54   &   53.49   &    48.88      &    60.23   &    63.35 \\
  \rowcolor[gray]{.9} $~~$ + \alg     & \textbf{22.20}      &     \textbf{92.60}        &    \textbf{79.92}   &   \textbf{53.66}   &    \textbf{56.58}      &    \textbf{61.83}   &    \textbf{63.89}  \\ \cmidrule(lr){1-8}
  SigLIP          &   36.66      &     95.86        &    88.47   &   61.70   &    33.88      &    75.14   &    43.49 \\ 
  \rowcolor[gray]{.9} $~~$ + \alg     &   \textbf{39.48}      &     \textbf{96.06}        &    \textbf{88.50}   &   \textbf{63.30}   &    \textbf{34.30}      &    \textbf{75.22}   &    \textbf{44.24} \\ 
    \bottomrule
  
    \end{tabular}
    \caption{\textbf{Generalization of \alg across different VLM backbones.}
We report top-1 classification accuracy (\%) on seven datasets using CLIP, BLIP-2, and SigLIP, with and without \alg.
Bold indicates the better result for each backbone.}
    \label{tab:various_vlms}
\end{table}

\subsection{Ablation on Uncertainty Metric}
To validate our choice of \textit{entropy} as the uncertainty metric, we performed an ablation by replacing it with two other common metrics: Max Softmax probability and Energy Score~\citep{liu2020energy}. In Table~\ref{tab:uncertain_metric}, the results indicate that entropy is the most effective choice, achieving the highest accuracy on six out of seven benchmark datasets. The gain is particularly pronounced on DTD, where entropy surpasses the next best metric by a large margin. While Max Softmax yields a stronger result on EuroSAT, the consistently top-tier performance of entropy across a diverse range of benchmarks indicates its general robustness.
\begin{table}[t]
  \centering
  \scriptsize
  \begin{tabular}{lccccccc}
  \toprule
  \textbf{Metric}  &  Aircraft    &    Caltech101    &    Cars    &    DTD    &    EuroSAT    &    IN      &    IN-A \\ \cmidrule(lr){1-8}
  Max Softmax      &   27.69      &     85.76        &    59.25   &   48.05   &    \textbf{82.14}      &    64.54   &    55.89 \\
  Energy           &   25.74      &     94.04        &    67.57   &   44.74   &    61.07      &    70.70   &    60.42  \\ 
  \rowcolor[gray]{.9} \textbf{Entropy (Ours)}        &   \textbf{27.84}      &     \textbf{94.06}        &    \textbf{68.76}   &   \textbf{51.63}   &    79.61      &    \textbf{70.88}   &    \textbf{60.73} \\ 
  \bottomrule
  
  \end{tabular}
    \caption{\textbf{Performance with various uncertainty metrics.} Since entropy, maximum softmax probability, and energy score have different numerical scales, we use metric-specific uncertainty thresholds chosen to match the same query budget, while keeping \(\tau_r=0.85\), \(N=5\), and the 10\% label budget fixed.}
    \label{tab:uncertain_metric}
    \vspace{-0.4cm}
\end{table}

\subsection{Comparison with Online Prompt Tuning}
\label{app:online_prompt_tuning}

Table~\ref{tab:coop_full} provides the full results of the
label-matched comparison across 15 datasets. Replay-CoOp
extends Online-CoOp by reusing accumulated labeled samples
for subsequent prompt updates. Despite this additional label
reuse, \alg achieves the highest average accuracy (67.65\%),
outperforming both Online-CoOp (65.15\%) and Replay-CoOp
(65.20\%).

\begin{table*}[t]
    \centering
    \scriptsize
    \setlength{\tabcolsep}{1.5pt}
    \begin{tabular}{l*{16}{c}}
    \toprule
    & \multicolumn{10}{c}{Cross-domain}
    & \multicolumn{5}{c}{Out-of-distribution}
    & \\
    \cmidrule(lr){2-11}
    \cmidrule(lr){12-16}
    \textbf{Method}
    & Aircraft & Caltech & Cars & DTD & EuroSAT
    & Flowers & Food & Pets & SUN & UCF
    & IN & IN-A & IN-V2 & IN-R & IN-S
    & \textbf{Avg.} \\
    \midrule

    Zero-shot CLIP
    & 22.59 & 85.76 & 65.61 & 40.60 & 44.25
    & 64.07 & 82.68 & 83.81 & 63.68 & 66.30
    & 63.52 & 54.65 & 59.80 & 75.36 & 44.05
    & 61.12 \\

    Online-CoOp
    & 26.22 & 89.17 & 66.68 & 42.61 & 72.43
    & 70.77 & 83.22 & 86.29 & \textbf{67.74} & 69.13
    & \textbf{68.50} & 52.47 & 59.73 & 76.09 & 46.13
    & 65.15 \\

    Replay-CoOp (online)
    & \textbf{27.63} & 89.82 & \textbf{68.26}
    & \textbf{49.29} & 76.58
    & 76.05 & 83.40 & 82.61 & 64.50 & 72.20
    & 62.73 & 54.27 & \textbf{61.23} & 69.92 & 39.53
    & 65.20 \\

    \rowcolor[gray]{.9}
    CLIP + \alg
    & 27.33 & \textbf{90.31} & 67.77 & 47.94
    & \textbf{80.63}
    & \textbf{80.53} & \textbf{83.57}
    & \textbf{87.10} & 65.30 & \textbf{76.28}
    & 65.90 & \textbf{55.85} & 59.69
    & \textbf{76.33} & \textbf{50.29}
    & \textbf{67.65} \\

    \bottomrule
    \end{tabular}
    \caption{
    \textbf{Full results of label-matched comparisons with online CoOp baselines.}
    Top-1 accuracy (\%) across 15 datasets using CLIP-ViT-B/16.
    All label-using methods share identical queried labels and follow
    the future-only supervision protocol.
    Best results are in \textbf{bold}.
    }    
    \label{tab:coop_full}
    \vspace{-0.4cm}
\end{table*}

\section{Conventional Sample Selection}
\label{app:baselines}
To evaluate the sample-selection component of \alg, we compare it against three standard streaming query-selection baselines: Random, Coreset, and Entropy. All methods follow the same Active VLM-TTA setup, use the same labeling budget, and reuse returned labels through the same retrieval memory. They differ only in how they select samples for annotation. 

\myparagraph{Random.} This baseline selects samples for annotation uniformly at random until the labeling budget is exhausted.

\myparagraph{Coreset~\citep{sener2018active}.} 
This baseline prioritizes samples that are far from the current labeled memory in the VLM embedding space. For a test sample $x_t$, we compute its maximum cosine similarity to the labeled memory and query samples with low similarity:
\[
S_{\rm core}(x_t) = - \max_{(x_i,y_i)\in D_l(t)} \cos(f_{\rm img}(x_t), f_{\rm img}(x_i)).
\]
Samples with larger \(S_{\rm core}(x_t)\) are prioritized for annotation.

\myparagraph{Entropy~\citep{holub2008entropy}.} 
This baseline prioritizes samples with high predictive uncertainty. Let \(q_t\) denote
the base TTA probability vector for \(x_t\). The selection score is
\[
S_{\rm ent}(x_t)=H(q_t)=-\sum_{c\in C}q_{t,c}\log q_{t,c}.
\]
Samples with larger \(S_{\rm ent}(x_t)\) are prioritized for annotation under the same labeling budget. Entropy is used only for query selection; retrieval-augmented correction still follows the same memory-based correction rule used by \alg.

\section{Societal Impact}
\label{app:impact}
Active VLM-TTA may improve the robustness of vision-language systems deployed in changing real-world environments by selectively incorporating limited human supervision, which could reduce repeated failures under distribution shift and make adaptation more label-efficient. Such improvements may be beneficial in applications where robustness matters, but the same paradigm also introduces risks: annotation errors or human bias can propagate through reused feedback, and limited labeling budgets may cause adaptation quality to differ across classes, scenarios, or user groups. As a result, deploying Active VLM-TTA in real-world systems requires careful attention to feedback quality, selection policy, and potential disparities in who benefits from adaptation.

\section{Licenses and Terms of Use for Existing Assets}
\label{app:licenses}

We use publicly available pretrained models, baseline implementations, and benchmark datasets as shown in Table~\ref{tab:license}.
All assets are used for research evaluation, and we do not redistribute third-party datasets.

\begin{table}[t]
\centering
\small
\begin{tabular}{llll}
\toprule
Asset & Type & Citation & License / Terms \\
\midrule
CLIP & Model/code & \citep{radford2021learning} & MIT License \\
BLIP-2 & Model/code & \citep{li2023blip} & BSD-3-Clause / model-card terms \\
SigLIP & Model & \citep{zhai2023sigmoid} & Apache-2.0 \\
TPT & Baseline code & \citep{shu2022tpt} & MIT License \\
C-TPT & Baseline code & \citep{yoon2024ctpt} & MIT License \\
TDA & Baseline code & \citep{karmanov2024efficient} & MIT License \\
CoOp & Baseline code & \citep{zhou2022learning} & MIT License \\
ImageNet & Dataset & \citep{deng2009imagenet} & ImageNet terms of access \\
ImageNet-A & Dataset & \citep{hendrycks2021nae} & MIT License \\
ImageNet-V2 & Dataset & \citep{recht2019imagenet} & MIT License \\
ImageNet-R & Dataset & \citep{hendrycks2021many} & MIT License \\
ImageNet-Sketch & Dataset & \citep{wang2019learning} & MIT License \\
FGVC-Aircraft & Dataset & \citep{maji2013fine} & Non-commercial research use \\
Caltech101 & Dataset & \citep{fei2004learning} & CC BY 4.0 \\
Stanford Cars & Dataset & \citep{krause20133d} & Research/ImageNet-style terms \\
DTD & Dataset & \citep{cimpoi2014describing} & Research use \\
EuroSAT & Dataset & \citep{helber2019eurosat} & MIT License; Sentinel data terms \\
Flowers102 & Dataset & \citep{nilsback2008automated} & Research use \\
Food101 & Dataset & \citep{bossard2014food} & Scientific fair-use / Foodspotting terms \\
Oxford Pets & Dataset & \citep{parkhi2012cats} & CC BY-SA 4.0 \\
SUN397 & Dataset & \citep{xiao2010sun} & Academic/research use \\
UCF101 & Dataset & \citep{soomro2012ucf101} & Research/fair-use style terms \\
\bottomrule
\end{tabular}
\caption{\textbf{Existing assets used in our experiments and their licenses or terms of use.}}
\label{tab:license}
\end{table}

%% file: checklist.tex
\section*{NeurIPS Paper Checklist}

\begin{enumerate}

\item {\bf Claims}
    \item[] Question: Do the main claims made in the abstract and introduction accurately reflect the paper's contributions and scope?
    \item[] Answer: \answerYes{} 
    \item[] Justification: The abstract and introduction accurately describe the paper’s main contributions: a streaming Active VLM-TTA setting with explicit future-only label supervision, the proposed training-free \alg framework, and empirical improvements under the stated budgeted protocol. These claims are supported by the method in Section~\ref{sec:method}, the main results in Section~\ref{sec:exp}, and additional analyses in Appendix~\ref{app:further_analyses}.
    \item[] Guidelines:
    \begin{itemize}
        \item The answer \answerNA{} means that the abstract and introduction do not include the claims made in the paper.
        \item The abstract and/or introduction should clearly state the claims made, including the contributions made in the paper and important assumptions and limitations. A \answerNo{} or \answerNA{} answer to this question will not be perceived well by the reviewers. 
        \item The claims made should match theoretical and experimental results, and reflect how much the results can be expected to generalize to other settings. 
        \item It is fine to include aspirational goals as motivation as long as it is clear that these goals are not attained by the paper. 
    \end{itemize}

\item {\bf Limitations}
    \item[] Question: Does the paper discuss the limitations of the work performed by the authors?
    \item[] Answer: \answerYes{} 
    \item[] Justification: The paper explicitly discusses limitations in Section~\ref{sec:conclusion}.
    \item[] Guidelines:
    \begin{itemize}
        \item The answer \answerNA{} means that the paper has no limitation while the answer \answerNo{} means that the paper has limitations, but those are not discussed in the paper. 
        \item The authors are encouraged to create a separate ``Limitations'' section in their paper.
        \item The paper should point out any strong assumptions and how robust the results are to violations of these assumptions (e.g., independence assumptions, noiseless settings, model well-specification, asymptotic approximations only holding locally). The authors should reflect on how these assumptions might be violated in practice and what the implications would be.
        \item The authors should reflect on the scope of the claims made, e.g., if the approach was only tested on a few datasets or with a few runs. In general, empirical results often depend on implicit assumptions, which should be articulated.
        \item The authors should reflect on the factors that influence the performance of the approach. For example, a facial recognition algorithm may perform poorly when image resolution is low or images are taken in low lighting. Or a speech-to-text system might not be used reliably to provide closed captions for online lectures because it fails to handle technical jargon.
        \item The authors should discuss the computational efficiency of the proposed algorithms and how they scale with dataset size.
        \item If applicable, the authors should discuss possible limitations of their approach to address problems of privacy and fairness.
        \item While the authors might fear that complete honesty about limitations might be used by reviewers as grounds for rejection, a worse outcome might be that reviewers discover limitations that aren't acknowledged in the paper. The authors should use their best judgment and recognize that individual actions in favor of transparency play an important role in developing norms that preserve the integrity of the community. Reviewers will be specifically instructed to not penalize honesty concerning limitations.
    \end{itemize}

\item {\bf Theory assumptions and proofs}
    \item[] Question: For each theoretical result, does the paper provide the full set of assumptions and a complete (and correct) proof?
    \item[] Answer: \answerNA{} 
    \item[] Justification: The paper does not make formal theoretical claims. 
    \item[] Guidelines:
    \begin{itemize}
        \item The answer \answerNA{} means that the paper does not include theoretical results. 
        \item All the theorems, formulas, and proofs in the paper should be numbered and cross-referenced.
        \item All assumptions should be clearly stated or referenced in the statement of any theorems.
        \item The proofs can either appear in the main paper or the supplemental material, but if they appear in the supplemental material, the authors are encouraged to provide a short proof sketch to provide intuition. 
        \item Inversely, any informal proof provided in the core of the paper should be complemented by formal proofs provided in appendix or supplemental material.
        \item Theorems and Lemmas that the proof relies upon should be properly referenced. 
    \end{itemize}

    \item {\bf Experimental result reproducibility}
    \item[] Question: Does the paper fully disclose all the information needed to reproduce the main experimental results of the paper to the extent that it affects the main claims and/or conclusions of the paper (regardless of whether the code and data are provided or not)?
    \item[] Answer: \answerYes{} 
    \item[] Justification: The paper provides the datasets, baselines, algorithm description, hyperparameters, evaluation protocol, and implementation details needed to reproduce the main results. These are described in Section~\ref{sec:exp}, Appendix~\ref{app:alg}, and Appendix~\ref{app:exp}, including Algorithm 1, dataset lists, thresholds, label budget, and model settings.
    \item[] Guidelines:
    \begin{itemize}
        \item The answer \answerNA{} means that the paper does not include experiments.
        \item If the paper includes experiments, a \answerNo{} answer to this question will not be perceived well by the reviewers: Making the paper reproducible is important, regardless of whether the code and data are provided or not.
        \item If the contribution is a dataset and\slash or model, the authors should describe the steps taken to make their results reproducible or verifiable. 
        \item Depending on the contribution, reproducibility can be accomplished in various ways. For example, if the contribution is a novel architecture, describing the architecture fully might suffice, or if the contribution is a specific model and empirical evaluation, it may be necessary to either make it possible for others to replicate the model with the same dataset, or provide access to the model. In general. releasing code and data is often one good way to accomplish this, but reproducibility can also be provided via detailed instructions for how to replicate the results, access to a hosted model (e.g., in the case of a large language model), releasing of a model checkpoint, or other means that are appropriate to the research performed.
        \item While NeurIPS does not require releasing code, the conference does require all submissions to provide some reasonable avenue for reproducibility, which may depend on the nature of the contribution. For example
        \begin{enumerate}
            \item If the contribution is primarily a new algorithm, the paper should make it clear how to reproduce that algorithm.
            \item If the contribution is primarily a new model architecture, the paper should describe the architecture clearly and fully.
            \item If the contribution is a new model (e.g., a large language model), then there should either be a way to access this model for reproducing the results or a way to reproduce the model (e.g., with an open-source dataset or instructions for how to construct the dataset).
            \item We recognize that reproducibility may be tricky in some cases, in which case authors are welcome to describe the particular way they provide for reproducibility. In the case of closed-source models, it may be that access to the model is limited in some way (e.g., to registered users), but it should be possible for other researchers to have some path to reproducing or verifying the results.
        \end{enumerate}
    \end{itemize}

\item {\bf Open access to data and code}
    \item[] Question: Does the paper provide open access to the data and code, with sufficient instructions to faithfully reproduce the main experimental results, as described in supplemental material?
    \item[] Answer: \answerYes{} 
    \item[] Justification: The paper provides detailed experimental settings, including datasets, baselines, and implementation details. The source code and instructions for reproducing the experiments are publicly available at \url{https://github.com/kaist-dmlab/ACTOR}.
    \item[] Guidelines:
    \begin{itemize}
        \item The answer \answerNA{} means that paper does not include experiments requiring code.
        \item Please see the NeurIPS code and data submission guidelines (\url{https://neurips.cc/public/guides/CodeSubmissionPolicy}) for more details.
        \item While we encourage the release of code and data, we understand that this might not be possible, so \answerNo{} is an acceptable answer. Papers cannot be rejected simply for not including code, unless this is central to the contribution (e.g., for a new open-source benchmark).
        \item The instructions should contain the exact command and environment needed to run to reproduce the results. See the NeurIPS code and data submission guidelines (\url{https://neurips.cc/public/guides/CodeSubmissionPolicy}) for more details.
        \item The authors should provide instructions on data access and preparation, including how to access the raw data, preprocessed data, intermediate data, and generated data, etc.
        \item The authors should provide scripts to reproduce all experimental results for the new proposed method and baselines. If only a subset of experiments are reproducible, they should state which ones are omitted from the script and why.
        \item At submission time, to preserve anonymity, the authors should release anonymized versions (if applicable).
        \item Providing as much information as possible in supplemental material (appended to the paper) is recommended, but including URLs to data and code is permitted.
    \end{itemize}

\item {\bf Experimental setting/details}
    \item[] Question: Does the paper specify all the training and test details (e.g., data splits, hyperparameters, how they were chosen, type of optimizer) necessary to understand the results?
    \item[] Answer: \answerYes{} 
    \item[] Justification: The paper specifies the datasets, baselines, model family, batch size, thresholds, initial labeled-sample count, budget, evaluation metric, and hardware setting. Additional algorithmic details and benchmark descriptions are provided in Appendix~\ref{app:alg} and Appendix~\ref{app:exp}.
    \item[] Guidelines:
    \begin{itemize}
        \item The answer \answerNA{} means that the paper does not include experiments.
        \item The experimental setting should be presented in the core of the paper to a level of detail that is necessary to appreciate the results and make sense of them.
        \item The full details can be provided either with the code, in appendix, or as supplemental material.
    \end{itemize}

\item {\bf Experiment statistical significance}
    \item[] Question: Does the paper report error bars suitably and correctly defined or other appropriate information about the statistical significance of the experiments?
    \item[] Answer: \answerYes{} 
    \item[] Justification: The main experimental tables report mean accuracy with standard deviations across three runs for the adaptive methods, and the implementation details explicitly state that all experiments were conducted three times and averaged with standard deviation.
    \item[] Guidelines:
    \begin{itemize}
        \item The answer \answerNA{} means that the paper does not include experiments.
        \item The authors should answer \answerYes{} if the results are accompanied by error bars, confidence intervals, or statistical significance tests, at least for the experiments that support the main claims of the paper.
        \item The factors of variability that the error bars are capturing should be clearly stated (for example, train/test split, initialization, random drawing of some parameter, or overall run with given experimental conditions).
        \item The method for calculating the error bars should be explained (closed form formula, call to a library function, bootstrap, etc.)
        \item The assumptions made should be given (e.g., Normally distributed errors).
        \item It should be clear whether the error bar is the standard deviation or the standard error of the mean.
        \item It is OK to report 1-sigma error bars, but one should state it. The authors should preferably report a 2-sigma error bar than state that they have a 96\% CI, if the hypothesis of Normality of errors is not verified.
        \item For asymmetric distributions, the authors should be careful not to show in tables or figures symmetric error bars that would yield results that are out of range (e.g., negative error rates).
        \item If error bars are reported in tables or plots, the authors should explain in the text how they were calculated and reference the corresponding figures or tables in the text.
    \end{itemize}

\item {\bf Experiments compute resources}
    \item[] Question: For each experiment, does the paper provide sufficient information on the computer resources (type of compute workers, memory, time of execution) needed to reproduce the experiments?
    \item[] Answer: \answerYes{} 
    \item[] Justification: The paper described the hardware spec (See Appendix~\ref{app:implementation_detail}), and reported the results of per-sample inference cost (See Table~\ref{tab:time_cost}).
    \item[] Guidelines:
    \begin{itemize}
        \item The answer \answerNA{} means that the paper does not include experiments.
        \item The paper should indicate the type of compute workers CPU or GPU, internal cluster, or cloud provider, including relevant memory and storage.
        \item The paper should provide the amount of compute required for each of the individual experimental runs as well as estimate the total compute. 
        \item The paper should disclose whether the full research project required more compute than the experiments reported in the paper (e.g., preliminary or failed experiments that didn't make it into the paper). 
    \end{itemize}
    
\item {\bf Code of ethics}
    \item[] Question: Does the research conducted in the paper conform, in every respect, with the NeurIPS Code of Ethics \url{https://neurips.cc/public/EthicsGuidelines}?
    \item[] Answer: \answerYes{} 
    \item[] Justification: To the best of our knowledge, the research complies with the NeurIPS Code of Ethics. The work uses standard benchmark datasets and model evaluation procedures, does not involve deceptive data collection or unsafe intervention, and does not report experiments involving protected or vulnerable human populations.
    \item[] Guidelines:
    \begin{itemize}
        \item The answer \answerNA{} means that the authors have not reviewed the NeurIPS Code of Ethics.
        \item If the authors answer \answerNo, they should explain the special circumstances that require a deviation from the Code of Ethics.
        \item The authors should make sure to preserve anonymity (e.g., if there is a special consideration due to laws or regulations in their jurisdiction).
    \end{itemize}

\item {\bf Broader impacts}
    \item[] Question: Does the paper discuss both potential positive societal impacts and negative societal impacts of the work performed?
    \item[] Answer: \answerYes{} 
    \item[] Justification: The paper discusses both potential benefits and risks of active test-time adaptation in Appendix~\ref{app:impact}.
    \item[] Guidelines:
    \begin{itemize}
        \item The answer \answerNA{} means that there is no societal impact of the work performed.
        \item If the authors answer \answerNA{} or \answerNo, they should explain why their work has no societal impact or why the paper does not address societal impact.
        \item Examples of negative societal impacts include potential malicious or unintended uses (e.g., disinformation, generating fake profiles, surveillance), fairness considerations (e.g., deployment of technologies that could make decisions that unfairly impact specific groups), privacy considerations, and security considerations.
        \item The conference expects that many papers will be foundational research and not tied to particular applications, let alone deployments. However, if there is a direct path to any negative applications, the authors should point it out. For example, it is legitimate to point out that an improvement in the quality of generative models could be used to generate Deepfakes for disinformation. On the other hand, it is not needed to point out that a generic algorithm for optimizing neural networks could enable people to train models that generate Deepfakes faster.
        \item The authors should consider possible harms that could arise when the technology is being used as intended and functioning correctly, harms that could arise when the technology is being used as intended but gives incorrect results, and harms following from (intentional or unintentional) misuse of the technology.
        \item If there are negative societal impacts, the authors could also discuss possible mitigation strategies (e.g., gated release of models, providing defenses in addition to attacks, mechanisms for monitoring misuse, mechanisms to monitor how a system learns from feedback over time, improving the efficiency and accessibility of ML).
    \end{itemize}
    
\item {\bf Safeguards}
    \item[] Question: Does the paper describe safeguards that have been put in place for responsible release of data or models that have a high risk for misuse (e.g., pre-trained language models, image generators, or scraped datasets)?
    \item[] Answer: \answerNA{} 
    \item[] Justification: The paper does not release a high-risk generative model, scraped dataset, or other asset that would require special safeguards for responsible release. The work studies adaptation methods on existing benchmark datasets and pretrained vision-language models.
    \item[] Guidelines:
    \begin{itemize}
        \item The answer \answerNA{} means that the paper poses no such risks.
        \item Released models that have a high risk for misuse or dual-use should be released with necessary safeguards to allow for controlled use of the model, for example by requiring that users adhere to usage guidelines or restrictions to access the model or implementing safety filters. 
        \item Datasets that have been scraped from the Internet could pose safety risks. The authors should describe how they avoided releasing unsafe images.
        \item We recognize that providing effective safeguards is challenging, and many papers do not require this, but we encourage authors to take this into account and make a best faith effort.
    \end{itemize}

\item {\bf Licenses for existing assets}
    \item[] Question: Are the creators or original owners of assets (e.g., code, data, models), used in the paper, properly credited and are the license and terms of use explicitly mentioned and properly respected?
    \item[] Answer: \answerYes{}
    \item[] Justification: All existing assets used in the paper, including pretrained models, baseline codebases, and benchmark datasets, are cited and their licenses or terms of use are explicitly listed in Appendix~\ref{app:licenses}. We use these assets only for research evaluation and do not redistribute third-party datasets.
    \item[] Guidelines:
        \begin{itemize}
        \item The answer \answerNA{} means that the paper does not use existing assets.
        \item The authors should cite the original paper that produced the code package or dataset.
        \item The authors should state which version of the asset is used and, if possible, include a URL.
        \item The name of the license (e.g., CC-BY 4.0) should be included for each asset.
        \item For scraped data from a particular source (e.g., website), the copyright and terms of service of that source should be provided.
        \item If assets are released, the license, copyright information, and terms of use in the package should be provided. For popular datasets, \url{paperswithcode.com/datasets} has curated licenses for some datasets. Their licensing guide can help determine the license of a dataset.
        \item For existing datasets that are re-packaged, both the original license and the license of the derived asset (if it has changed) should be provided.
        \item If this information is not available online, the authors are encouraged to reach out to the asset's creators.
    \end{itemize}

\item {\bf New assets}
    \item[] Question: Are new assets introduced in the paper well documented and is the documentation provided alongside the assets?
    \item[] Answer: \answerYes{} 
    \item[] Justification: We introduce \alg, a new algorithmic framework for \textbf{Active VLM-TTA}. The implementation and documentation are publicly available through our GitHub repository at \url{https://github.com/kaist-dmlab/ACTOR}.
    \item[] Guidelines:
    \begin{itemize}
        \item The answer \answerNA{} means that the paper does not release new assets.
        \item Researchers should communicate the details of the dataset\slash code\slash model as part of their submissions via structured templates. This includes details about training, license, limitations, etc. 
        \item The paper should discuss whether and how consent was obtained from people whose asset is used.
        \item At submission time, remember to anonymize your assets (if applicable). You can either create an anonymized URL or include an anonymized zip file.
    \end{itemize}

\item {\bf Crowdsourcing and research with human subjects}
    \item[] Question: For crowdsourcing experiments and research with human subjects, does the paper include the full text of instructions given to participants and screenshots, if applicable, as well as details about compensation (if any)? 
    \item[] Answer: \answerNA{} 
    \item[] Justification: The paper does not conduct crowdsourcing experiments or human-subject studies. 
    \item[] Guidelines:
    \begin{itemize}
        \item The answer \answerNA{} means that the paper does not involve crowdsourcing nor research with human subjects.
        \item Including this information in the supplemental material is fine, but if the main contribution of the paper involves human subjects, then as much detail as possible should be included in the main paper. 
        \item According to the NeurIPS Code of Ethics, workers involved in data collection, curation, or other labor should be paid at least the minimum wage in the country of the data collector. 
    \end{itemize}

\item {\bf Institutional review board (IRB) approvals or equivalent for research with human subjects}
    \item[] Question: Does the paper describe potential risks incurred by study participants, whether such risks were disclosed to the subjects, and whether Institutional Review Board (IRB) approvals (or an equivalent approval/review based on the requirements of your country or institution) were obtained?
    \item[] Answer: \answerNA{} 
    \item[] Justification: The paper does not involve human-subject data collection or human-subject experimentation requiring IRB review.
    \item[] Guidelines:
    \begin{itemize}
        \item The answer \answerNA{} means that the paper does not involve crowdsourcing nor research with human subjects.
        \item Depending on the country in which research is conducted, IRB approval (or equivalent) may be required for any human subjects research. If you obtained IRB approval, you should clearly state this in the paper. 
        \item We recognize that the procedures for this may vary significantly between institutions and locations, and we expect authors to adhere to the NeurIPS Code of Ethics and the guidelines for their institution. 
        \item For initial submissions, do not include any information that would break anonymity (if applicable), such as the institution conducting the review.
    \end{itemize}

\item {\bf Declaration of LLM usage}
    \item[] Question: Does the paper describe the usage of LLMs if it is an important, original, or non-standard component of the core methods in this research? Note that if the LLM is used only for writing, editing, or formatting purposes and does \emph{not} impact the core methodology, scientific rigor, or originality of the research, declaration is not required.
    \item[] Answer: \answerNA{} 
    \item[] Justification: In this paper, the LLM is used only for writing and editing.
    \item[] Guidelines:
    \begin{itemize}
        \item The answer \answerNA{} means that the core method development in this research does not involve LLMs as any important, original, or non-standard components.
        \item Please refer to our LLM policy in the NeurIPS handbook for what should or should not be described.
    \end{itemize}

\end{enumerate}

%% file: custom.bib
@String(ARXIV = {ArXiv Preprint})

@String(ICML =  {Proceedings of International Conference on Machine Learning})

@String(CVPR  = {Proceedings of IEEE/CVF Conference on Computer Vision and Pattern Recognition})

@String(CVPRW = {Proceedings of IEEE/CVF Conference on Computer Vision and Pattern Recognition Workshop})

@String(ICCV  = {Proceedings of IEEE/CVF International Conference on Computer Vision})

@String(ICCVW = {Proceedings of IEEE/CVF International Conference on Computer Vision Workshop})

@String(ECCV  = {Proceedings of European Conference on Computer Vision})

@String(NEURIPS  = {Proceedings of Advances in Neural Information Processing Systems})

@String(ICLR  = {Proceedings of International Conference on Learning Representations})

@String(ECML = {Proceedings of European Conference on Machine Learning})

@String(ICVGIP = {Proceedings of Indian Conference on Computer Vision, Graphics \& Image Processing})

@String(IJCV  = {International Journal of Computer Vision})

@inproceedings{radford2021learning,
  title={Learning transferable visual models from natural language supervision},
  author={Radford, Alec and Kim, Jong Wook and Hallacy, Chris and Ramesh, Aditya and Goh, Gabriel and Agarwal, Sandhini and Sastry, Girish and Askell, Amanda and Mishkin, Pamela and Clark, Jack and others},
  booktitle=ICML,
  pages={8748--8763},
  year={2021}
}

@article{bai2023qwen,
  title={Qwen technical report},
  author={Bai, Jinze and Bai, Shuai and Chu, Yunfei and Cui, Zeyu and Dang, Kai and Deng, Xiaodong and Fan, Yang and Ge, Wenbin and Han, Yu and Huang, Fei and others},
  journal=ARXIV,
  year={2023}
}

@inproceedings{li2022blip,
  title={Blip: Bootstrapping language-image pre-training for unified vision-language understanding and generation},
  author={Li, Junnan and Li, Dongxu and Xiong, Caiming and Hoi, Steven},
  booktitle=ICML,
  pages={12888--12900},
  year={2022},
  organization={PMLR}
}

@inproceedings{li2023blip,
  title={Blip-2: Bootstrapping language-image pre-training with frozen image encoders and large language models},
  author={Li, Junnan and Li, Dongxu and Savarese, Silvio and Hoi, Steven},
  booktitle=ICML,
  pages={19730--19742},
  year={2023},
  organization={PMLR}
}

@article{zhou2022learning,
  title={Learning to prompt for vision-language models},
  author={Zhou, Kaiyang and Yang, Jingkang and Loy, Chen Change and Liu, Ziwei},
  journal=IJCV,
  volume={130},
  number={9},
  pages={2337--2348},
  year={2022},
  publisher={Springer}
}

@inproceedings{zhou2022conditional,
  title={Conditional prompt learning for vision-language models},
  author={Zhou, Kaiyang and Yang, Jingkang and Loy, Chen Change and Liu, Ziwei},
  booktitle=CVPR,
  pages={16816--16825},
  year={2022}
}

@inproceedings{khattak2023maple,
  title={Maple: Multi-modal prompt learning},
  author={Khattak, Muhammad Uzair and Rasheed, Hanoona and Maaz, Muhammad and Khan, Salman and Khan, Fahad Shahbaz},
  booktitle=CVPR,
  pages={19113--19122},
  year={2023}
}

@inproceedings{karmanov2024efficient,
  title={Efficient Test-Time Adaptation of Vision-Language Models},
  author={Karmanov, Adilbek and Guan, Dayan and Lu, Shijian and El Saddik, Abdulmotaleb and Xing, Eric},
  booktitle=CVPR,
  pages={14162--14171},
  year={2024}
}

@inproceedings{shu2022tpt,
  title={{Test-time Prompt Tuning for Zero-shot Generalization in Vision-Language Models}},
  author={Shu, Manli and Nie, Weili and Huang, De-An and Yu, Zhiding and Goldstein, Tom and Anandkumar, Anima and Xiao, Chaowei},
  booktitle=NEURIPS,
  pages={14274--14289},
  year={2022}
}

@inproceedings{feng2023difftpt,
  title={{Diverse Data Augmentation with Diffusions for Effective Test-time Prompt Tuning}},
  author={Feng, Chun-Mei and Yu, Kai and Liu, Yong and Khan, Salman and Zuo, Wangmeng},
  booktitle=ICCV,
  pages={2704--2714},
  year={2023}
}

@inproceedings{yoon2024ctpt,
  title={{C-TPT: Calibrated Test-Time Prompt Tuning for Vision-Language Models via Text Feature Dispersion}},
  author={Yoon, Hee Suk and Yoon, Eunseop and Tee, Joshua Tian Jin and Hasegawa-Johnson, Mark and Li, Yingzhen and Yoo, Chang D},
  booktitle=ICLR,
  year={2024}
}

@inproceedings{zhao2023rlcf,
  title={{Test-time Adaptation with CLIP Reward for Zero-shot Generalization in Vision-Language Models}},
  author={Zhao, Shuai and Wang, Xiaohan and Zhu, Linchao and Yang, Yi},
  booktitle=ICLR,
  year={2024}
}

@article{yao2021filip,
  title={Filip: Fine-grained interactive language-image pre-training},
  author={Yao, Lewei and Huang, Runhui and Hou, Lu and Lu, Guansong and Niu, Minzhe and Xu, Hang and Liang, Xiaodan and Li, Zhenguo and Jiang, Xin and Xu, Chunjing},
  journal=ARXIV,
  year={2021}
}

@inproceedings{liu2023visual,
  title={Visual instruction tuning},
  author={Liu, Haotian and Li, Chunyuan and Wu, Qingyang and Lee, Yong Jae},
  booktitle=NEURIPS,
  volume={36},
  year={2023}
}

@article{gao2023clip,
  title={Clip-adapter: Better vision-language models with feature adapters},
  author={Gao, Peng and Geng, Shijie and Zhang, Renrui and Ma, Teli and Fang, Rongyao and Zhang, Yongfeng and Li, Hongsheng and Qiao, Yu},
  journal=IJCV,
  pages={1--15},
  year={2023},
  publisher={Springer}
}

@inproceedings{jia2021scaling,
  title={Scaling up visual and vision-language representation learning with noisy text supervision},
  author={Jia, Chao and Yang, Yinfei and Xia, Ye and Chen, Yi-Ting and Parekh, Zarana and Pham, Hieu and Le, Quoc and Sung, Yun-Hsuan and Li, Zhen and Duerig, Tom},
  booktitle=ICML,
  pages={4904--4916},
  year={2021}
}

@inproceedings{zhai2022lit,
  title={Lit: Zero-shot transfer with locked-image text tuning},
  author={Zhai, Xiaohua and Wang, Xiao and Mustafa, Basil and Steiner, Andreas and Keysers, Daniel and Kolesnikov, Alexander and Beyer, Lucas},
  booktitle=CVPR,
  pages={18123--18133},
  year={2022}
}

@article{ash2019deep,
  title={Deep batch active learning by diverse, uncertain gradient lower bounds},
  author={Ash, Jordan T and Zhang, Chicheng and Krishnamurthy, Akshay and Langford, John and Agarwal, Alekh},
  journal=ARXIV,
  year={2019}
}

@techreport{settles2009active,
  title={Active learning literature survey},
  author={Settles, Burr},
  year={2009},
  institution={Department of Computer Sciences, University of Wisconsin-Madison}
}

@article{ren2021survey,
  title={A survey of deep active learning},
  author={Ren, Pengzhen and Xiao, Yun and Chang, Xiaojun and Huang, Po-Yao and Li, Zhihui and Gupta, Brij B and Chen, Xiaojiang and Wang, Xin},
  journal={ACM Computing Surveys (CSUR)},
  volume={54},
  number={9},
  pages={1--40},
  year={2021},
  publisher={ACM New York, NY}
}

@inproceedings{holub2008entropy,
  title={Entropy-based active learning for object recognition},
  author={Holub, Alex and Perona, Pietro and Burl, Michael C},
  booktitle=CVPRW,
  pages={1--8},
  year={2008}
}

@inproceedings{parvaneh2022active,
  title={Active learning by feature mixing},
  author={Parvaneh, Amin and Abbasnejad, Ehsan and Teney, Damien and Haffari, Gholamreza Reza and Van Den Hengel, Anton and Shi, Javen Qinfeng},
  booktitle=CVPR,
  pages={12237--12246},
  year={2022}
}

@inproceedings{sener2018active,
  title={Active Learning for Convolutional Neural Networks: A Core-Set Approach},
  author={Sener, Ozan and Savarese, Silvio},
  booktitle=ICLR,
  year={2018}
}

@article{helber2019eurosat,
  title={Eurosat: A novel dataset and deep learning benchmark for land use and land cover classification},
  author={Helber, Patrick and Bischke, Benjamin and Dengel, Andreas and Borth, Damian},
  journal={IEEE Journal of Selected Topics in Applied Earth Observations and Remote Sensing},
  volume={12},
  number={7},
  pages={2217--2226},
  year={2019},
  publisher={IEEE}
}

@inproceedings{parkhi2012cats,
  title={Cats and dogs},
  author={Parkhi, Omkar M and Vedaldi, Andrea and Zisserman, Andrew and Jawahar, CV},
  booktitle=CVPR,
  pages={3498--3505},
  year={2012}
}

@inproceedings{cimpoi2014describing,
  title={Describing textures in the wild},
  author={Cimpoi, Mircea and Maji, Subhransu and Kokkinos, Iasonas and Mohamed, Sammy and Vedaldi, Andrea},
  booktitle=CVPR,
  pages={3606--3613},
  year={2014}
}

@inproceedings{fei2004learning,
  title={Learning generative visual models from few training examples: An incremental bayesian approach tested on 101 object categories},
  author={Fei-Fei, Li and Fergus, Rob and Perona, Pietro},
  booktitle=CVPRW,
  pages={178--178},
  year={2004}
}

@inproceedings{nilsback2008automated,
  title={Automated flower classification over a large number of classes},
  author={Nilsback, Maria-Elena and Zisserman, Andrew},
  booktitle= ICVGIP,
  pages={722--729},
  year={2008}
}

@inproceedings{krause20133d,
  title={3d object representations for fine-grained categorization},
  author={Krause, Jonathan and Stark, Michael and Deng, Jia and Fei-Fei, Li},
  booktitle=ICCVW,
  pages={554--561},
  year={2013}
}

@article{maji2013fine,
  title={Fine-grained visual classification of aircraft},
  author={Maji, Subhransu and Rahtu, Esa and Kannala, Juho and Blaschko, Matthew and Vedaldi, Andrea},
  journal=ARXIV,
  year={2013}
}

@inproceedings{lewis1994heterogeneous,
  title={Heterogeneous uncertainty sampling for supervised learning},
  author={Lewis, David D and Catlett, Jason},
  booktitle=ICML,
  pages={148--156},
  year={1994},
}

@inproceedings{roth2006margin,
  title={Margin-based active learning for structured output spaces},
  author={Roth, Dan and Small, Kevin},
  booktitle=ECML,
  pages={413--424},
  year={2006}
}

@inproceedings{deng2009imagenet,
  title={Imagenet: A large-scale hierarchical image database},
  author={Deng, Jia and Dong, Wei and Socher, Richard and Li, Li-Jia and Li, Kai and Fei-Fei, Li},
  booktitle=CVPR,
  pages={248--255},
  year={2009}
}

@inproceedings{bossard2014food,
  title={Food-101--mining discriminative components with random forests},
  author={Bossard, Lukas and Guillaumin, Matthieu and Van Gool, Luc},
  booktitle=ECCV,
  pages={446--461},
  year={2014}
}

@inproceedings{xiao2010sun,
  title={Sun database: Large-scale scene recognition from abbey to zoo},
  author={Xiao, Jianxiong and Hays, James and Ehinger, Krista A and Oliva, Aude and Torralba, Antonio},
  booktitle=CVPR,
  pages={3485--3492},
  year={2010}
}

@article{soomro2012ucf101,
  title={UCF101: A dataset of 101 human actions classes from videos in the wild},
  author={Soomro, Khurram and Zamir, Amir Roshan and Shah, Mubarak},
  journal=ARXIV,
  year={2012}
}

@inproceedings{hendrycks2021nae,
  title={Natural Adversarial Examples},
  author={Dan Hendrycks and Kevin Zhao and Steven Basart and Jacob Steinhardt and Dawn Song},
  booktitle=CVPR,
  year={2021}
}

@inproceedings{recht2019imagenet,
  title={Do imagenet classifiers generalize to imagenet?},
  author={Recht, Benjamin and Roelofs, Rebecca and Schmidt, Ludwig and Shankar, Vaishaal},
  booktitle=ICML,
  pages={5389--5400},
  year={2019},
  organization={PMLR}
}

@inproceedings{hendrycks2021many,
  title={The many faces of robustness: A critical analysis of out-of-distribution generalization},
  author={Hendrycks, Dan and Basart, Steven and Mu, Norman and Kadavath, Saurav and Wang, Frank and Dorundo, Evan and Desai, Rahul and Zhu, Tyler and Parajuli, Samyak and Guo, Mike and others},
  booktitle=ICCV,
  pages={8340--8349},
  year={2021}
}

@article{wang2019learning,
  title={Learning robust global representations by penalizing local predictive power},
  author={Wang, Haohan and Ge, Songwei and Lipton, Zachary and Xing, Eric P},
  journal=NEURIPS,
  volume={32},
  year={2019}
}

@inproceedings{gui2024active,
  title={Active Test-Time Adaptation: Theoretical Analyses and An Algorithm},
  author={Gui, Shurui and Li, Xiner and Ji, Shuiwang},
  booktitle=ICLR,
  year={2024}
}

@article{qin2021active,
  title={Active learning with extreme learning machine for online imbalanced multiclass classification},
  author={Qin, Jiongming and Wang, Cong and Zou, Qinhong and Sun, Yubin and Chen, Bin},
  journal={Knowledge-Based Systems},
  volume={231},
  pages={107385},
  year={2021},
  publisher={Elsevier}
}

@inproceedings{desalvo2021online,
  title={Online active learning with surrogate loss functions},
  author={DeSalvo, Giulia and Gentile, Claudio and Thune, Tobias Sommer},
  booktitle=NEURIPS,
  volume={34},
  pages={22877--22889},
  year={2021}
}

@inproceedings{bang2024active,
  title={Active prompt learning in vision language models},
  author={Bang, Jihwan and Ahn, Sumyeong and Lee, Jae-Gil},
  booktitle=CVPR,
  pages={27004--27014},
  year={2024}
}

@inproceedings{zhang2022tip,
  title={Tip-adapter: Training-free adaption of clip for few-shot classification},
  author={Zhang, Renrui and Zhang, Wei and Fang, Rongyao and Gao, Peng and Li, Kunchang and Dai, Jifeng and Qiao, Yu and Li, Hongsheng},
  booktitle=ECCV,
  pages={493--510},
  year={2022},
  organization={Springer}
}

@inproceedings{zhai2023sigmoid,
  title={Sigmoid loss for language image pre-training},
  author={Zhai, Xiaohua and Mustafa, Basil and Kolesnikov, Alexander and Beyer, Lucas},
  booktitle=ICCV,
  pages={11975--11986},
  year={2023}
}

@article{liu2020energy,
  title={Energy-based out-of-distribution detection},
  author={Liu, Weitang and Wang, Xiaoyun and Owens, John and Li, Yixuan},
  journal=NEURIPS,
  volume={33},
  pages={21464--21475},
  year={2020}
}

@article{bai2025qwen2,
  title={Qwen2.5-vl technical report},
  author={Bai, Shuai and Chen, Keqin and Liu, Xuejing and Wang, Jialin and Ge, Wenbin and Song, Sibo and Dang, Kai and Wang, Peng and Wang, Shijie and Tang, Jun and others},
  journal=ARXIV,
  year={2025}
}

@inproceedings{lee2025ra,
  title={Ra-tta: Retrieval-augmented test-time adaptation for vision-language models},
  author={Lee, Youngjun and Kim, Doyoung and Kang, Junhyeok and Bang, Jihwan and Song, Hwanjun and Lee, Jae-Gil},
  booktitle=ICLR,
  year={2025}
}

@inproceedings{zancato2023train,
  title={Train/test-time adaptation with retrieval},
  author={Zancato, Luca and Achille, Alessandro and Liu, Tian Yu and Trager, Matthew and Perera, Pramuditha and Soatto, Stefano},
  booktitle=CVPR,
  pages={15911--15921},
  year={2023}
}
